\documentclass[10pt,twocolumn,letterpaper]{article}
\usepackage{iccv}
\usepackage{times}
\usepackage{blindtext}
\usepackage{multicol}
\usepackage{graphicx}
\usepackage{comment}
\usepackage{amsmath,amssymb}
\usepackage{color}
\usepackage{epsfig}
\usepackage{booktabs}
\usepackage{multirow}
\usepackage{array}
\usepackage{subfigure}
\usepackage{setspace}
\usepackage{wrapfig}
\usepackage{arydshln}
\usepackage{algorithm} 
\usepackage{algorithmic}
\usepackage{autobreak}
\usepackage{ragged2e}
\usepackage{scalerel}
\usepackage[title]{appendix}
\usepackage{amsthm}

\newtheorem{corollary}{Corollary}

\newtheorem{assumption}{Assumption}

\usepackage[pagebackref=true,breaklinks=true,letterpaper=true,colorlinks,bookmarks=false]{hyperref}

\iccvfinalcopy % *** Uncomment this line for the final submission

\ificcvfinal\fi

\begin{document}
	
	%%%%%%%%% TITLE
	\title{Amplitude-Phase Recombination: Rethinking Robustness of Convolutional Neural Networks in Frequency Domain}
	
	\author{Guangyao Chen\textsuperscript{1}, Peixi Peng\textsuperscript{1,3$\ast$}, Li Ma\textsuperscript{1,3}, Jia Li\textsuperscript{2,3}, Lin Du\textsuperscript{4}, Yonghong Tian\textsuperscript{1,3}\thanks{Corresponding author} \\
		\small \textsuperscript{1}Department of Computer Science and Technology, Peking University \ \  \textsuperscript{3}Peng Cheng Laborotory \\
		\small \textsuperscript{2}State Key Laboratory of Virtual Reality Technology and Systems, SCSE, Beihang University \ \  \textsuperscript{4}AI Application Research Center, Huawei \\
		{\tt\small \{gy.chen, pxpeng, mali\_hp, yhtian\}@pku.edu.cn, jiali@buaa.edu.cn, dulin09@huawei.com}
		% Department of Computer Science and Technology, 
		% State Key Laboratory of Virtual Reality Technology and Systems, SCSE, 
		% \and
	}

	\maketitle
	\ificcvfinal\thispagestyle{empty}\fi
	%%%%%%%%% ABSTRACT
	\begin{abstract}
		Recently, the generalization behavior of Convolutional Neural Networks (CNN) is gradually transparent through explanation techniques with the frequency components decomposition. However, the importance of the phase spectrum of the image for a robust vision system is still ignored. In this paper, we notice that the CNN tends to converge at the local optimum which is closely related to the high-frequency components of the training images, while the amplitude spectrum is easily disturbed such as noises or common corruptions. In contrast, more empirical studies found that humans rely on more phase components to achieve robust recognition. This observation leads to more explanations of the CNN's generalization behaviors in both robustness to common perturbations and out-of-distribution detection, and motivates a new perspective on data augmentation designed by re-combing the phase spectrum of the current image and the amplitude spectrum of the distracter image. That is, the generated samples force the CNN to pay more attention to the structured information from phase components and keep robust to the variation of the amplitude. Experiments on several image datasets indicate that the proposed method achieves state-of-the-art performances on multiple generalizations and calibration tasks, including adaptability for common corruptions and surface variations, out-of-distribution detection, and adversarial attack. The code is released on \href{https://github.com/iCGY96/APR}{github/iCGY96/APR}.
		\vspace{-4mm}
	\end{abstract}
	
	%%%%%%%%% BODY TEXT
	\section{Introduction}

	In the past few years, deep learning has achieved even surpassed human-level performances in many image recognition/classification tasks \cite{he2015delving}.
	However, the unintuitive generalization behaviors of neural networks, such as the vulnerability towards adversarial examples \cite{goodfellow2014explaining}, common corruptions \cite{hendrycks2019augmix}, the overconfidence for out-of-distribution (OOD) \cite{hendrycks17baseline,hsu2020generalized,scheirer2012toward,chen_2020_ECCV,chen2021adversarial},  are still confused in the community.
	It also leads that current deep learning models depend on the ability of training data to faithfully represent the data encountered during deployment.
	
	\begin{figure}[!tb]
		\centering
		\setlength{\abovecaptionskip}{0.cm}
		\includegraphics[width=0.9\linewidth]{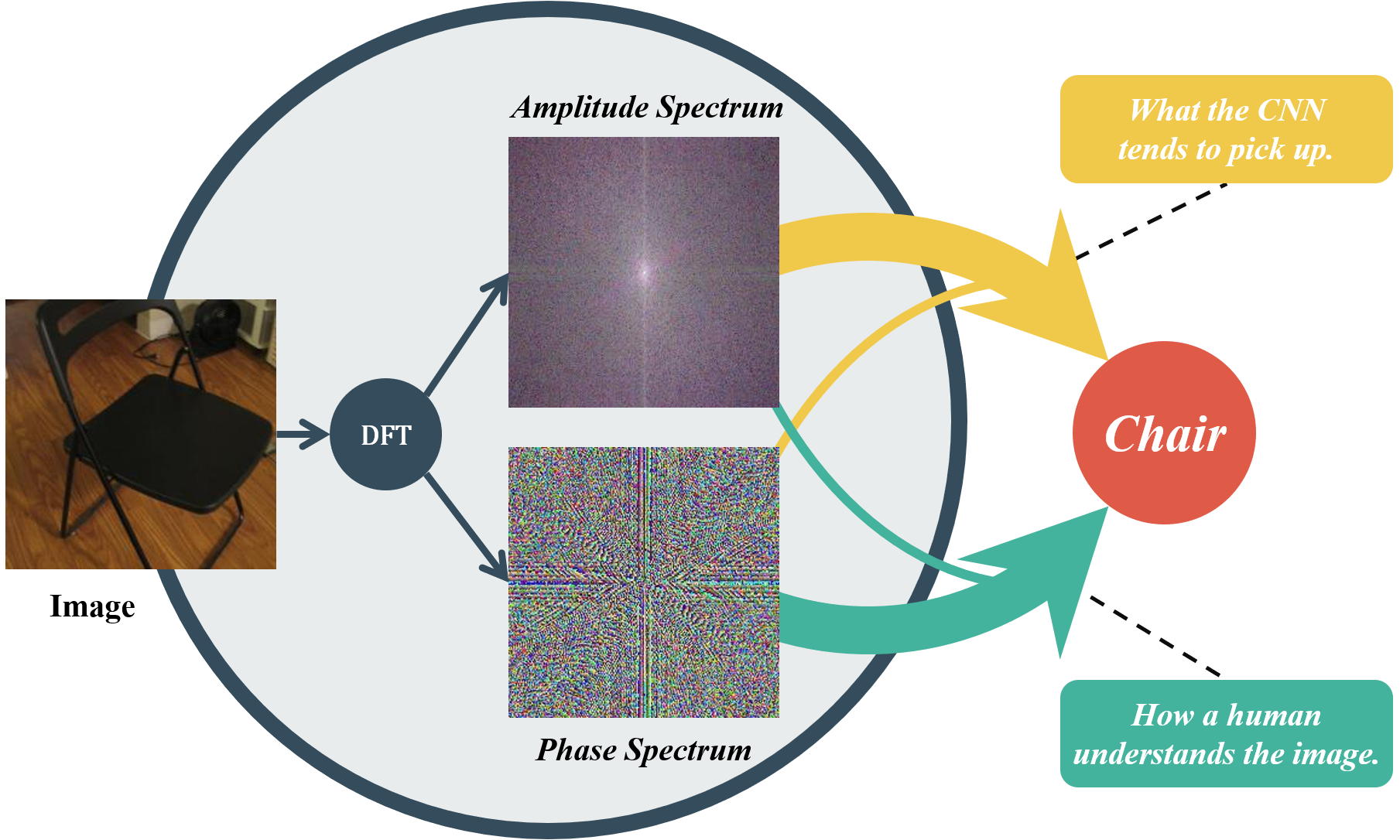}
		\caption{
			More empirical studies found that humans rely on more phase components to achieve robust recognition
			However, CNN without effective training restrictions tends to converge at the local optimum related to the amplitude spectrum of the image, leading to generalization behaviors counter-intuitive to humans.
		}
		\label{fig:motivation}
		\vspace{-4mm}
	\end{figure}
	
\begin{figure}[!tb]
		\centering
		\setlength{\abovecaptionskip}{0.cm}
		\includegraphics[width=0.9\linewidth]{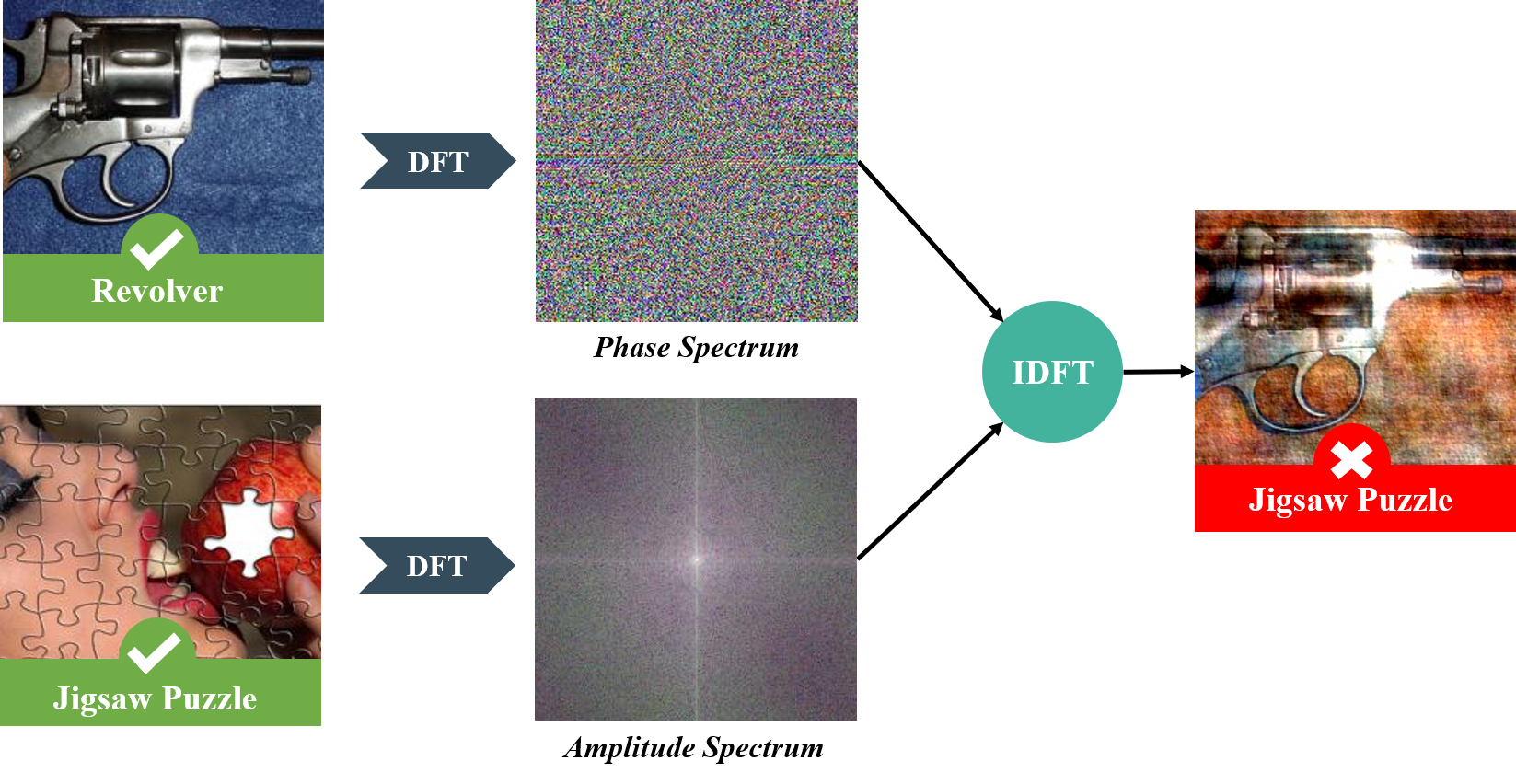}
		\caption{An example of the importance of phase spectrum to explain the counter-intuitive behavior of CNN. The recombined image with the phase spectrum of \emph{Revolver} and the amplitude spectrum of \emph{Jigsaw Puzzle} is recognized as \emph{Jigsaw Puzzle} by CNN. However, the human can still clearly recognize it as a \emph{Revolver}.}
		\label{fig:dft}
		\vspace{-4mm}
	\end{figure}

\begin{figure*}[!tb]
		\centering
		\setlength{\abovecaptionskip}{0.cm}
		\subfigure[In-distribution samples of airplane and frog]{
			\begin{minipage}[t]{0.47\linewidth}
				\centering
				\includegraphics[width=\linewidth]{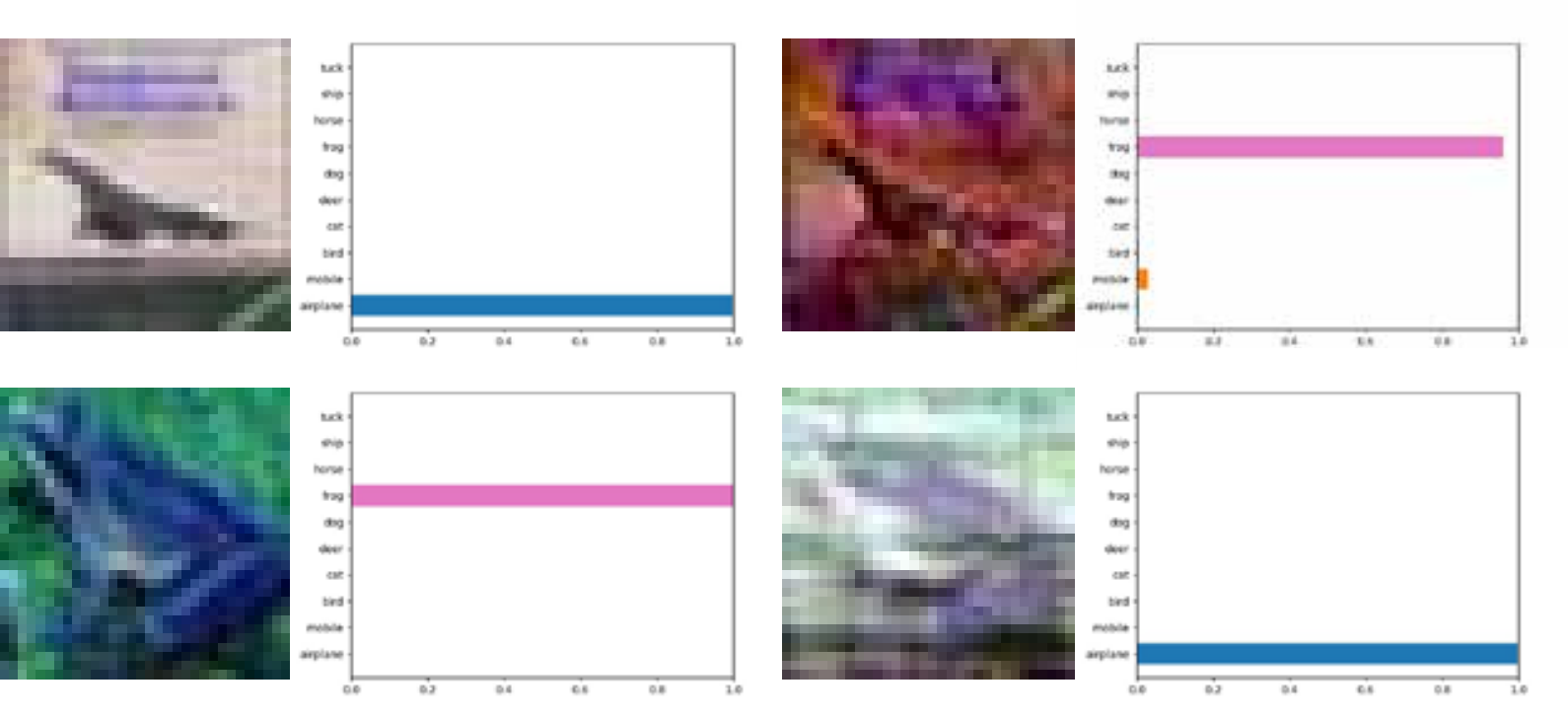}
				\label{fig:in1}
			\end{minipage}%
			
		}
		\subfigure[In-distribution samples of cat and bird]{
			\begin{minipage}[t]{0.47\linewidth}
				\centering
				\includegraphics[width=\linewidth]{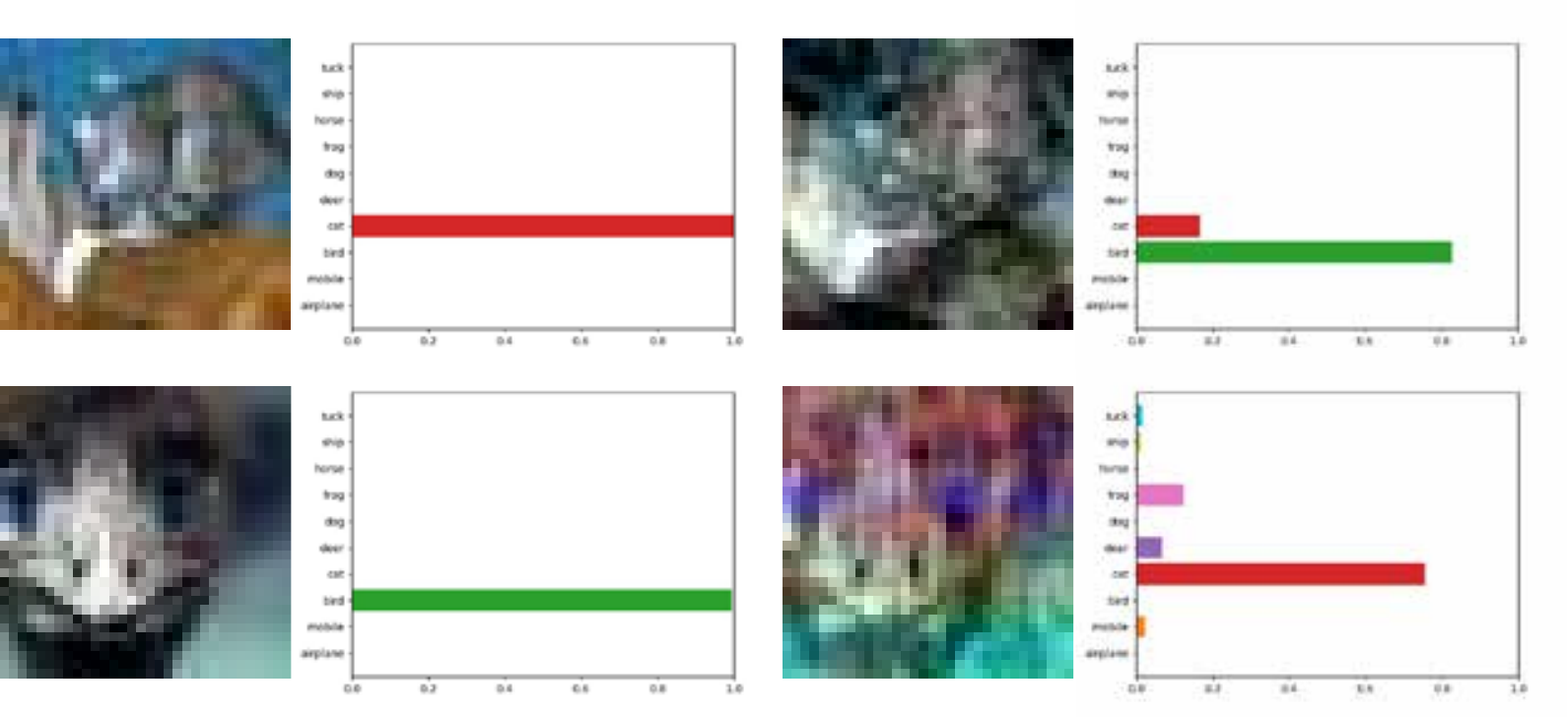}
				\label{fig:in2}
			\end{minipage}%
		}
		\subfigure[Out-of-distribution samples of $5$ and $6$]{
			\begin{minipage}[t]{0.47\linewidth}
				\centering
				\includegraphics[width=\linewidth]{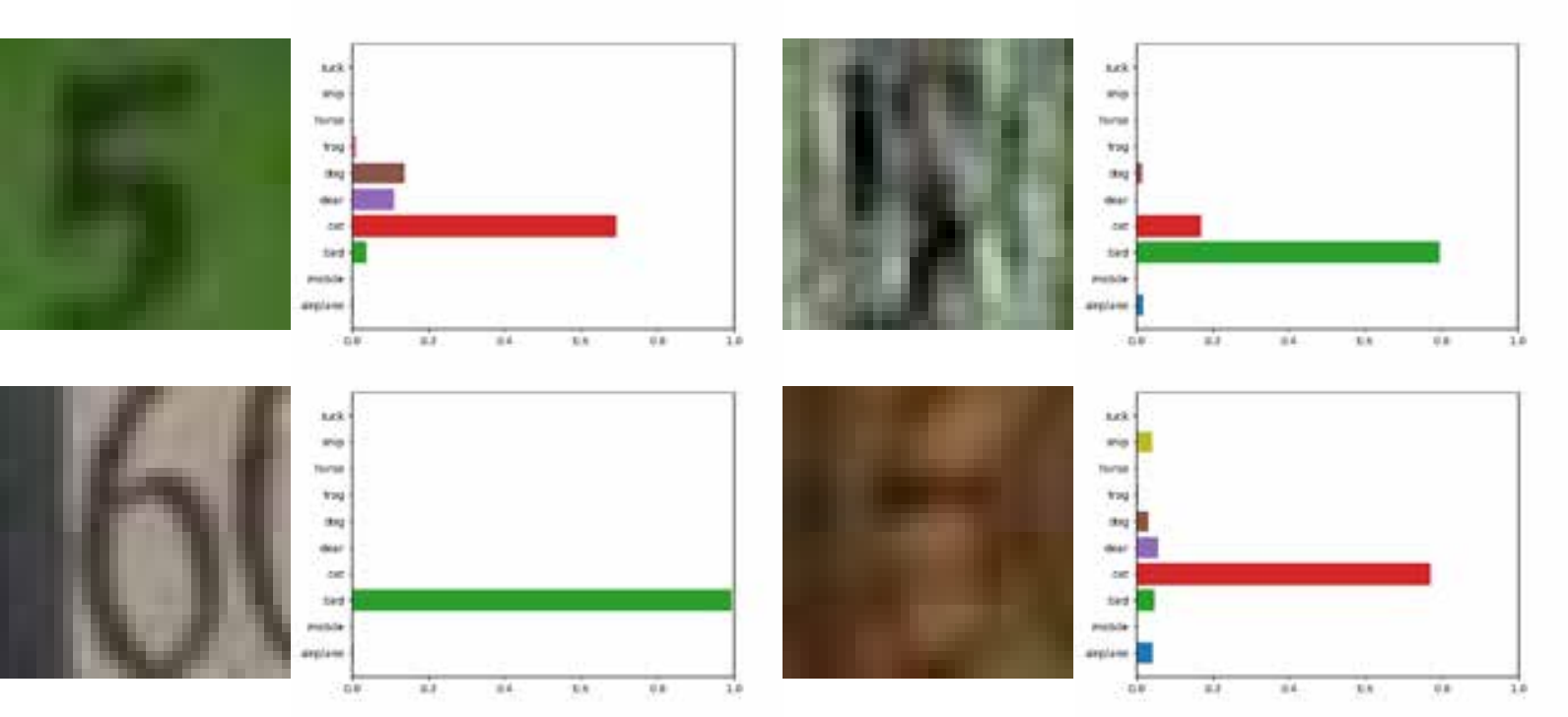}
				\label{fig:out1}
			\end{minipage}%
		}%
		\subfigure[Out-of-distribution samples of $2$ and $4$]{
			\begin{minipage}[t]{0.47\linewidth}
				\centering
				\includegraphics[width=\linewidth]{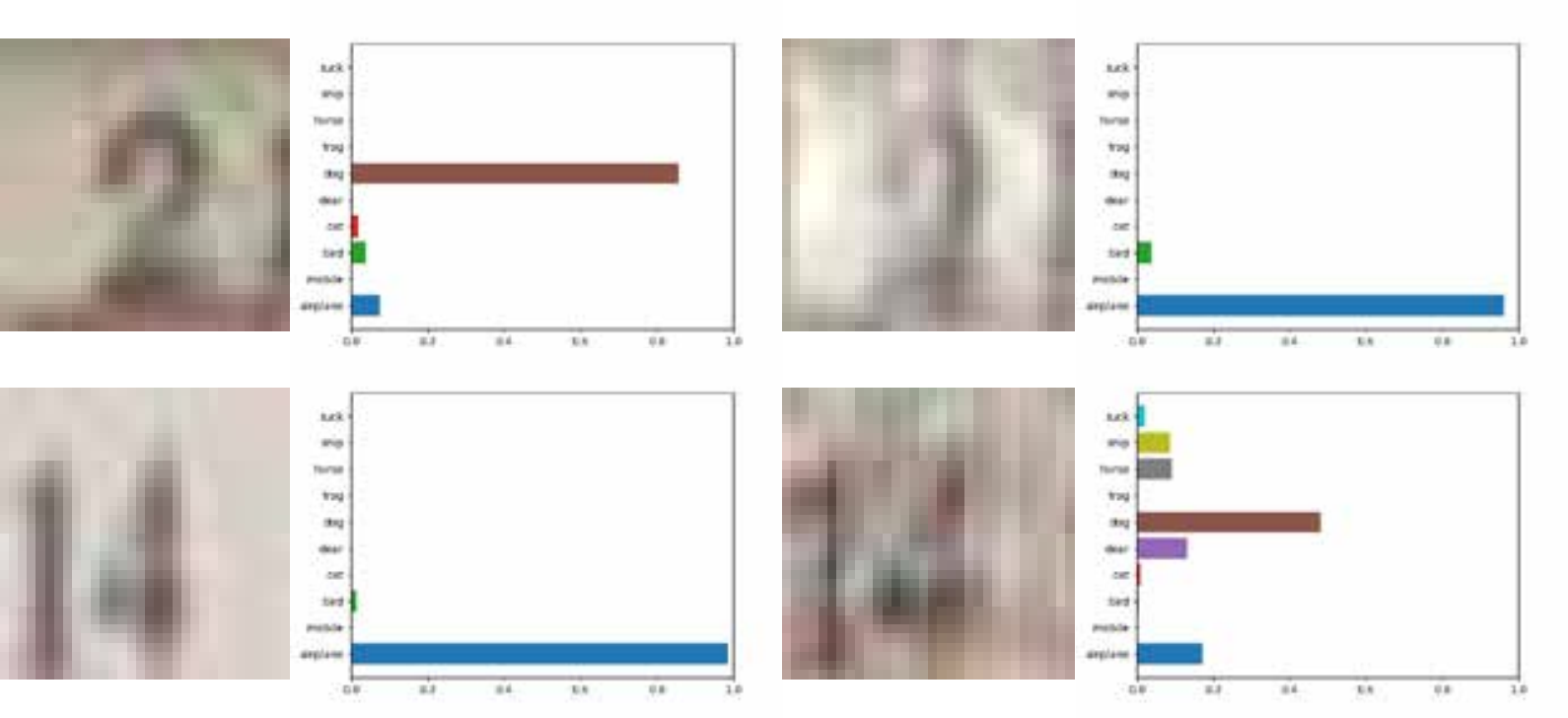}
				\label{fig:out2}
			\end{minipage}%
		}%
		\caption{Four pairs of testing samples selected from in-distribution CIFAR-10 \cite{krizhevsky2009learning} and OOD SVHN that help explain that CNN captures more amplitude spectrum than phase spectrum for classification: First, in (a) and (b), the model (All Convolutional Network) correctly predicts the original image ($1^{st}$ column in each panel), but the predicts are also exchanged after switching amplitude spectrum ($3^{rd}$ column in each panel) while the human eye can still give the correct category through the contour information. Secondly, the model is overconfident for the OOD samples in (c) and (d). Similarly, after the exchange of amplitude spectrum, the label with high confidence is also exchanged.}
		\label{fig:examples}
		\vspace{-4mm}
	\end{figure*}

	To explain the generalization behaviors of neural networks, many theoretical breakthroughs have been made progressively by different model or algorithm perspectives \cite{you2017scaling,schmidt2018adversarially,shamir2019simple}.
	Several works \cite{wang2020high,ilyas2019adversarial} investigate the generalization behaviors of Convolutional Neural Network (CNN) from a data perspective in the frequency domain, and demonstrate that CNN benefits from the high-frequency image components which are not perceivable to humans. Furthermore, a quantitative study is provided in Figure \ref{fig:AP_exp} to indicate that the predictions of CNN are more sensitive to the variation of the amplitude spectrum. 
	The above phenomena indicate that CNN tends to converge at the local optimum which is closely related to the high-frequency components of the training images.
	Although it is helpful when the test and training samples come from the identical distribution, yet the robustness of the CNN will be affected due of the amplitude spectrum is easily disturbed such as noises or common corruptions. 
	On the other hand, earlier empirical studies \cite{oppenheim1981importance,ghiglia1998two,guo2008spatio,li2015finding} indicate that humans rely on more the components related to the phase to recognize an object. 
	As is known, the human eye is much more robust than CNN, and this fact encourages us to rethink the influence of amplitude and phase on CNN's generalizability. 
	A visualized example is shown in Figure \ref{fig:dft} to validate the importance of phase spectrum in \cite{oppenheim1981importance} to explain one counter-intuitive behavior of CNN. 
	By replacing the amplitude spectrum of one \emph{Revolver} with the amplitude spectrum of one \emph{Jigsaw Puzzle}, the CNN classifies the fused image as \emph{Jigsaw Puzzle} while humans could still recognize it as \emph{Revolver}.
	In this example, the prediction outcomes of CNN are almost entirely determined by the amplitude spectrum of the image, which is barely perceivable to humans.
	On the other hand, even if the amplitude spectrum is replaced, the human is able to correctly recognize the identical object in the original picture.
	Moreover, we found that this phenomenon not only exists in training data (in-distribution) but also in OOD data as shown in Figure \ref{fig:examples}.
	In these images, after exchanging the amplitude spectrum, the prediction of CNN is also transformed with the label of the amplitude spectrum.
	However, humans could still observe the object structure of the original images in the converted images.
	
	Motivated by the powerful generalizability of the human, we argue that a robust CNN should be insensitive to the change of amplitude and pay more attention to the phase spectrum.
	To achieve this goal, a novel data augmentation method, called  Amplitude-Phase Recombination (APR), is proposed.
	The core of APR is to re-combine the phase spectrum of the current image and the amplitude spectrum of the distracter image to generate a new training sample, whose label is set to the current image.
	That is, the generated samples force the CNN to capture more structured information from phase components rather than amplitude.
	Specifically, the distracter image of the current image comes in two ways: other images and its augmentations generated by existing data augmentation methods such as rotate and random crop, namely APR  for the pair images (APR-P) and APR for the single image (APR-S) respectively.
	
	Extensive experiments on multiple generalizations and calibration tasks, including adaptability for common corruptions and surface variations, OOD detection, and adversarial attack, demonstrate the proposed APR outperforms the baselines by a large margin. 
	Meanwhile, it provides a uniform explanation to the texture bias hypothesis \cite{geirhos2018imagenet} and the behaviors of both robustness to common perturbations and the overconfidence of OOD by the CNN’s over-dependence on the amplitude spectrum. 
	That is, the various common perturbations change the high-frequency amplitude components significantly, while has little influence on the components related to the phase spectrum. 
	Hence, the attack sample could confuse the CNN but is easily recognized by humans. 
	On the other hand, the OOD samples often exhibit totally different image structures but may share some similarities in the high-frequency amplitude components, which makes the CNN hard to distinguish.
	% Moreover, it also provides much stronger evidence for the texture bias hypothesis \cite{geirhos2018imagenet}.
	
	Our main contributions are summarized as follows: 1) We propose that a robust CNN should be robust to the amplitude variance and pay more attention to the components related to the phase spectrum by a series of quantitative and qualitative analysis, 2) a novel data augmentation method APR is proposed to force the CNN pay more attention to the phase spectrum and achieves state-of-the-art performances on multiple generalizations and calibration tasks, including adaptability for common corruptions and surface variations, OOD detection, and adversarial attack, and 3) a unified explanation is provided to the behaviors of both robustness to common perturbations and the overconfidence of OOD by the CNN’s over-dependence on the amplitude spectrum.

	\section{Related Work}
	\label{sec:related}
	
	\textbf{Frequency-Based Explanation for CNN.}
	Recently, several works provide new insights into neural network behaviors from the aspects of the frequency domain. 
	\cite{wang2020high} shows that high-frequency components play significant roles in promoting CNN's accuracy, unlike human beings.
	Based on this observation, \cite{wang2020high} concludes that smoothing the CNN kernels helps to enforce the model to use features of low frequencies.
	\cite{guo2020low} proposes an adversarial attack only targeting the low-frequency components in an image, which shows that the model does utilize the features in the low-frequency domains for predictions instead of only learning from high-frequency components. 
	\cite{sharma2019effectiveness} demonstrates that state-of-the-art defenses are nearly as vulnerable as undefended models under low-frequency perturbations, which implies current defense techniques are only valid against adversarial attacks in the high-frequency domain. 
	On the other side, \cite{liu2021spatial} demonstrates that CNNs can capture extra implicit features of the phase spectrum which are beneficial to face forgery detection. 
	However, there are not works to give a qualitative study of the roles of amplitude and phase spectrums for the generalization behavior of CNN.
	
	\textbf{Data Augmentation.} 
	Data augmentation has been widely used to prevent deep neural networks from over-fitting to the training data \cite{bishop1995training}, and greatly improve generalization performance. 
	The majority of conventional augmentation methods generate new data by applying transformations depending on the data type or the target task \cite{cubuk2019autoaugment}. 
	\cite{zhang2017mixup} proposes \emph{mixup}, which linearly interpolates between two input data and utilizes the mixed data with the corresponding soft label for training. 
	Then, \emph{CutMix} \cite{yun2019cutmix} suggests a spatial copy and paste based mixup strategy on images. 
	AutoAugment \cite{cubuk2019autoaugment} is a learned augmentation method, where a group of augmentations is tuned to optimize performance on a downstream task.
	AugMix \cite{hendrycks2019augmix} helps models withstand unforeseen corruptions by simply mixing random augmentations.
	However, many methods substantially degrade accuracy on non-adversarial images \cite{raghunathan2019adversarial} or need adaptive and complex parameters to different tasks.
	
	\section{The Secret of CNN in the Frequency Domain}
	\label{sec:phase}
	
	\subsection{Qualitative Study on the Frequency Domain}
	
	Beyond the examples in Figure \ref{fig:dft} and \ref{fig:examples}, here more qualitative analyses are given to measure the contributions of amplitude and phase.
	Several experiments are conducted on CIFAR-10 \cite{krizhevsky2009learning} to evaluate the performances of the CNNs which are trained with the inversed images by various types of amplitude and phase spectra.
	For the image $x$, its frequency domain $\mathcal{F}_x$ is composed by amplitude $\mathcal{A}_x$ and phase $\mathcal{P}_x$ as:
	\begin{equation}\small
		\setlength{\abovedisplayskip}{3pt}
		\setlength{\belowdisplayskip}{3pt}
		\mathcal{F}_x = \mathcal{A}_x \otimes e^{i \cdot \mathcal{P}_x},
		\label{eqn:freq}
	\end{equation}
	where $\otimes$ indicates the element-wise multiplication of two matrices.
	Here, four types of amplitude spectra, $\mathcal{P}_x$, $\mathcal{P}_x^L$, $\mathcal{P}_x^I$, and $\mathcal{P}_x^H$ are combined with four types of amplitude spectra, including $\mathcal{A}_x$, $\mathcal{A}_x^L$, $\mathcal{A}_x^I$, and $\mathcal{A}_x^H$, respectively. 
	Here $\mathcal{A}_x^L$, $\mathcal{A}_x^I$, $\mathcal{A}_x^H$ and $\mathcal{P}_x^L$, $\mathcal{P}_x^I$, $\mathcal{P}_x^H$  represent the amplitude spectrum and phase of low-frequency, intermediate-frequency and high-frequency by low-pass $\mathcal{H}_l$, high-pass $\mathcal{H}_h$, and band-pass $\mathcal{H}_b$ filters, respectively. 
	Noted in Eq.\eqref{eqn:freq}, if one element of $\mathcal{A}_x$ is zero, then the corresponding element of $\mathcal{F}_x$ would be zero, and the phase spectrum $\mathcal{P}_x$ is not able to be considered. To alleviate the influence of this, we define the transfer function as:
	\begin{align*}\small
		\setlength{\abovedisplayskip}{3pt}
		\setlength{\belowdisplayskip}{3pt}
		\begin{split}
			\hat{z} = \left \{
			\begin{array}{ll}
				1,  & z = 0\\
				z,  & otherwise.
			\end{array}
			\right.
		\end{split}
	\end{align*}
	Finally, $\mathcal{P}_x$, $\hat{\mathcal{P}_x^L}$, $\hat{\mathcal{P}_x^I}$, and $\hat{\mathcal{P}_x^H}$ are combined with $\mathcal{A}_x$, $\hat{\mathcal{A}_x^L}$, $\hat{\mathcal{A}_x^I}$, and $\hat{\mathcal{A}_x^H}$, respectively.
	
	\begin{figure}[!tb]
		\centering
		\setlength{\abovecaptionskip}{0.cm}
		\includegraphics[width=0.8\linewidth]{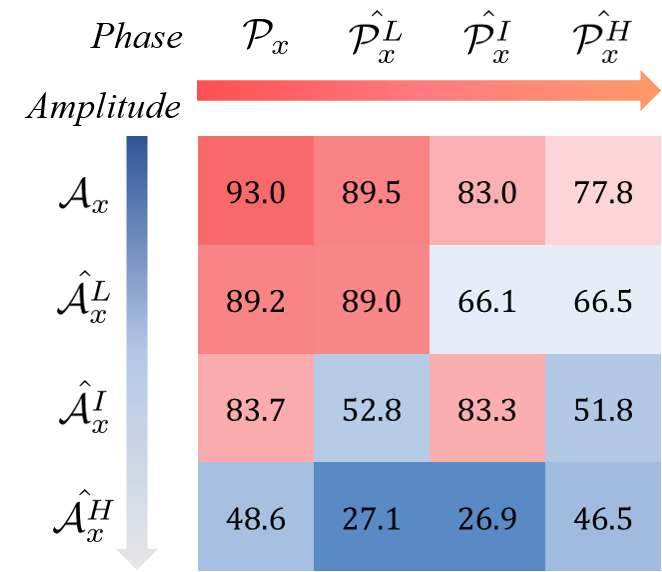}
		\caption{We test the classification power of CNNs trained with various combinations of the amplitude and phase spectrum.}
		\label{fig:AP_exp}
		\vspace{-4mm}
	\end{figure}
	
	For quantitative evaluation, we trained the ResNet-18 with the inversed images by the above each pair of amplitude and phase spectra:
	\begin{equation}\small
		\setlength{\abovedisplayskip}{3pt}
		\setlength{\belowdisplayskip}{3pt}
		\mathop{\arg\min}_{\theta}{ \ l(f(iDFT(\mathcal{A}_x \otimes e^{i \cdot \mathcal{P}_x}); \theta), y)},
	\end{equation}
	where $iDFT$ is the inverse Discrete Fourier Transform (DFT), and $f(\cdot)$ means the CNN model with the learnable parameters $\theta$.
	
	The test accuracies of the model trained by each pair are shown in Figure \ref{fig:AP_exp}. It is clear that the combination of phase and amplitude in the corresponding frequency domain achieves better performance in their various combinations, which indicates the CNN can capture effective information from both amplitude and phase spectrum. Moreover, when fixing the amplitude spectrum and phase spectrum respectively, the range of change without amplitude is larger than the case without phase according to the two directions of the arrow. It indicates that the convergence of the CNN more relies on the amplitude spectrum and neglects the phase spectrum.
	
\begin{figure}[!tb]
		\centering
		\setlength{\abovecaptionskip}{0.cm}
		\subfigure[Corrupted Samples]{
			\begin{minipage}[t]{0.48\linewidth}
				\centering
				\includegraphics[width=\linewidth]{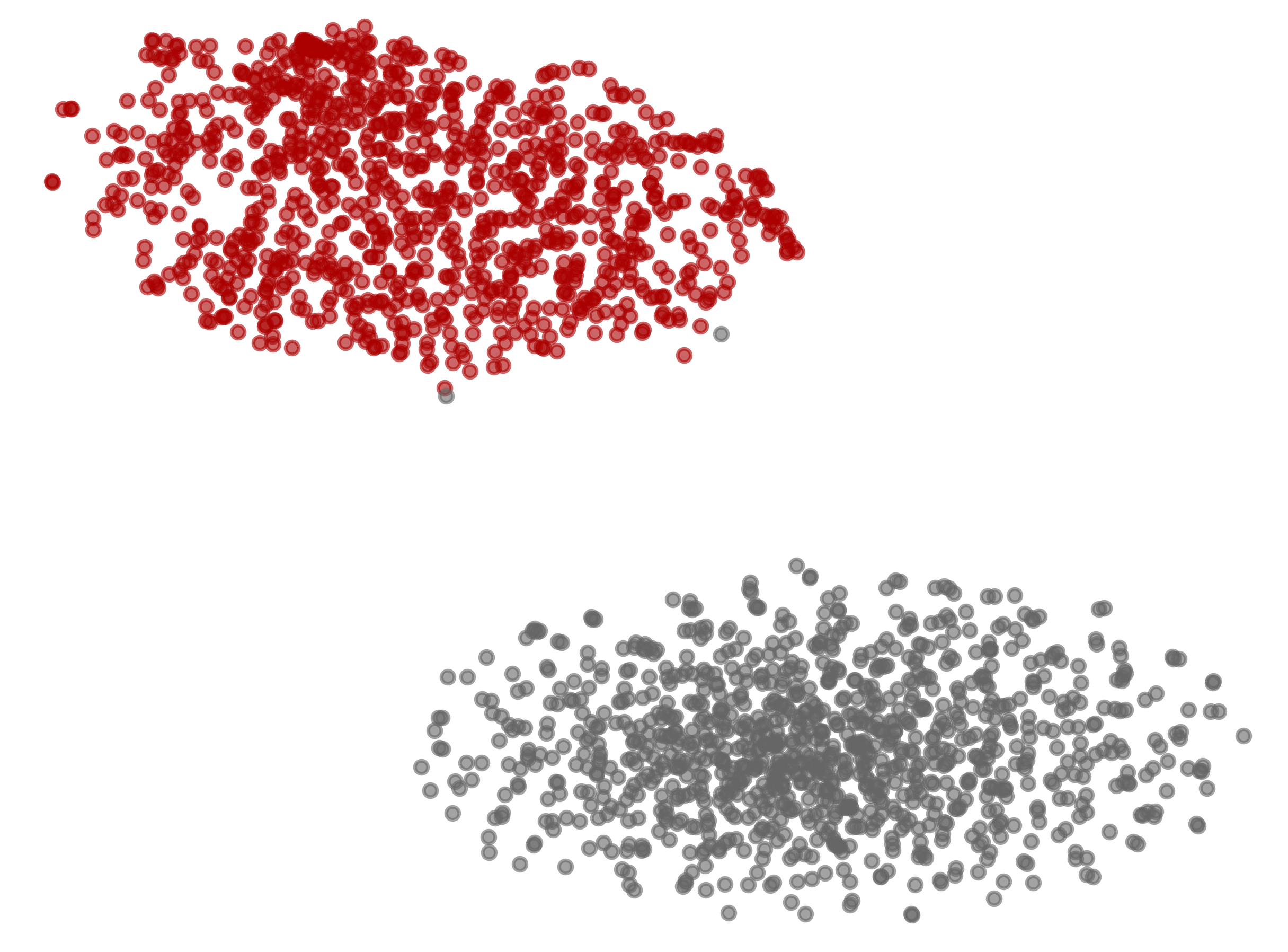}
				\label{fig:attck}
			\end{minipage}%
		}%
		\subfigure[OOD Samples]{
			\begin{minipage}[t]{0.48\linewidth}
				\centering
				\includegraphics[width=\linewidth]{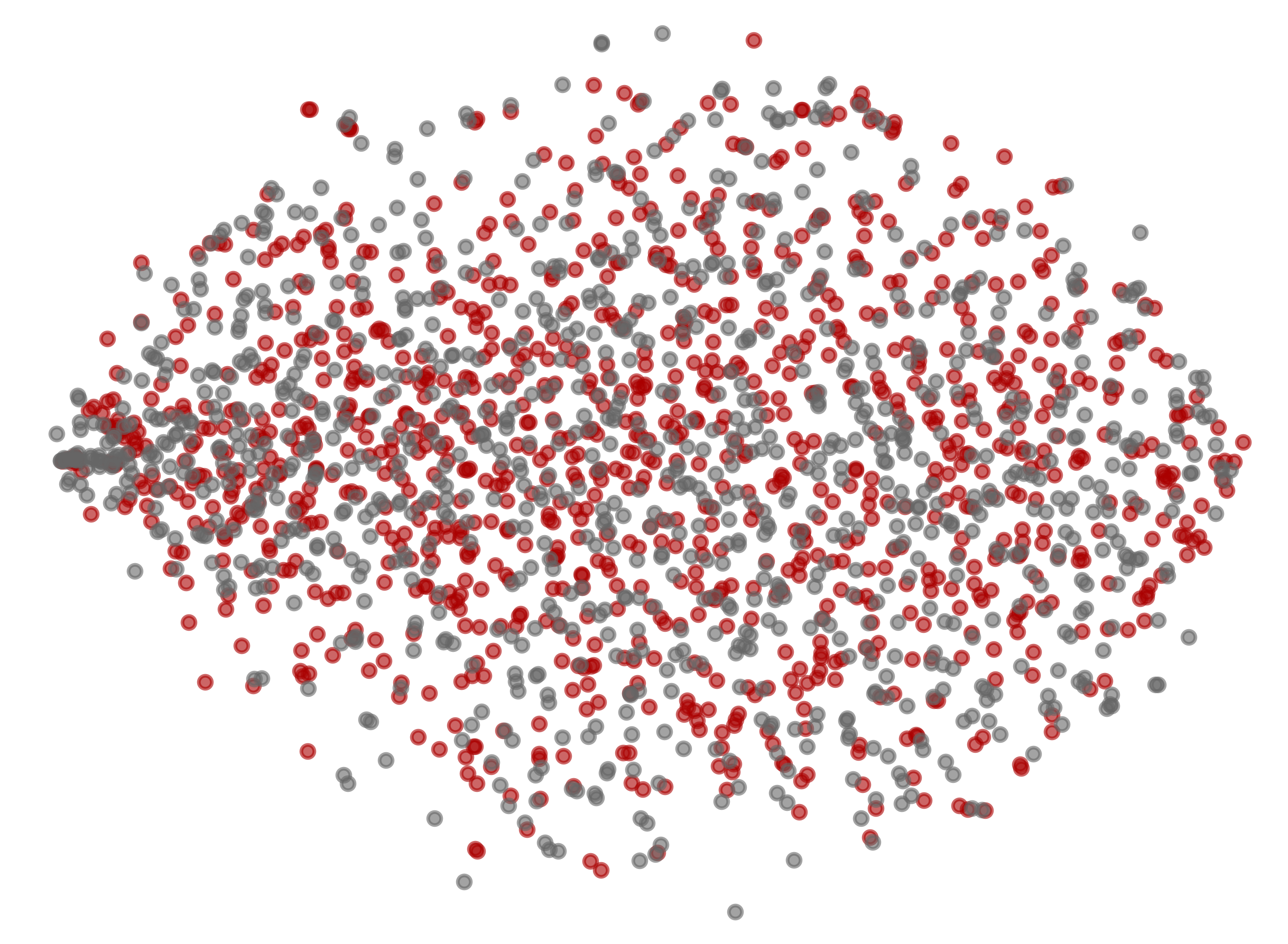}
				\label{fig:ood}
			\end{minipage}%
		}%
		\caption{The T-SNE \cite{van2008visualizing} distribution of the amplitude spectrum of high-frequency. Red represents the original image or in-distribution (ID) samples in CIFAR-10, and gray represents the corrupted samples from CIFAR-10 or OOD samples from CIFAR-100.}
		\label{fig:assum}
		\vspace{-4mm}
	\end{figure}

	Furthermore, we randomly select $1000$ samples from CIFAR-10.
	Firstly, we generate 1000 corrupted samples by Gaussian noise and show the distribution of corrupted samples and original samples as shown in Figure \ref{fig:attck}.
	We could observe the amplitude spectrum in high-frequency of two types of samples is so different while the corrupted sample is just added invisible noise.
	Hence, CNN would make the wrong prediction when the amplitude spectrum is changed.
	This is also consistent with the conclusion that CNN captured high-frequency information in \cite{wang2020high}.
	Therefore, we propose an assumption (referred to as A1) that presumes:
	\begin{assumption}
		\label{assumption:1}
		CNN without effective training restrictions tends to perceive more amplitude spectrum instead of the phase spectrum.
	\end{assumption}
	\noindent
	Then, we can formulate another formal statement for the robustness of CNN as:
	\begin{corollary}
		\label{corollary:1}
		With the assumption A1, there exists a sample $\langle x,y \rangle$ with its amplitude $\mathcal{A}_x$ and phase $\mathcal{P}_x$, that the model $f(\cdot)$ without effective training restrictions cannot predict robustly for $\hat{x} = iDFT((\mathcal{A}_x+\epsilon) \otimes e^{i \cdot \mathcal{P}_x})$ where $\epsilon$ is the upper bound of the perturbation allowed.
	\end{corollary}
	
	Secondly, we randomly select $1000$ OOD samples from CIFAR-100.
	As shown in Figure \ref{fig:ood}, it is not able to distinguish the amplitude spectrum in high-frequency of in-distribution and out-of-distribution, even these samples are from different categories.
	As a result, CNN would be overconfident for some distributions when similar amplitude information appears.
	Therefore, we first attempt to provide an assumption (referred to as A2) for the behaviors of the robustness to common perturbations and the overconfidence of OOD:
	\begin{assumption}
		\label{assumption:2}
		The behaviors of the sensitivity to common perturbations and the overconfidence of OOD may be all due to CNN’s over-dependence on the amplitude spectrum.
	\end{assumption}
	% We assume that "both of them are all due to an over-focus on the amplitude spectrum" (referred to as A2).
	\noindent
	Meanwhile, we can extend our main argument for OOD to a new formal statement:
	\begin{corollary}
		\label{corollary:2}
		With the assumptions A1 and A2, there exists a in-distribution sample $\langle x_1,y \rangle$ and an out-of-distribution sample $\langle x_2 \rangle$ with their amplitude $\mathcal{A}_{x_1}, \mathcal{A}_{x_2}$ and phase $\mathcal{P}_{x_1}, \mathcal{P}_{x_2}$, that the model without effective training restrictions would give a high confidence of the $y$ for $\hat{x} = iDFT(\mathcal{A}_{x_1} \otimes e^{i \cdot \mathcal{P}_{x_2}})$.
	\end{corollary}
	
	The proof is a direct outcome of the previous discussion and thus omitted. 
	The Corollary \ref{corollary:1} has been proved in previous works \cite{wang2020high,schmidt2018adversarially,shamir2019simple} and Corollary \ref{corollary:2} can also be verified empirically (e.g., in Figure \ref{fig:dft} and \ref{fig:examples}), therefore we can safely state that these two corollaries can serve as the alternative explanations to the generalization behavior of CNN.
	Meanwhile, we provide more examples for proof in Appendix.
	
	\subsection{The Role of the Phase Spectrum}
	Previous works \cite{oppenheim1981importance,ghiglia1998two} have shown many of the important features of a signal are preserved if only the phase spectrum is retained.
	Meanwhile, several works of image saliency \cite{guo2008spatio,li2015finding} shown the connection of phase spectrum with the fixation of the human visual system.
	Further, we wish to explore why this important information of the image is retained in the phase spectrum.
	Here, we reinterpret the concept of discrete Fourier transforms from the perspective of template-based contrast computation \cite{li2015finding}.
	
	Give a gray image $x$ with resolution $N \times N$, its complex-valued Fourier coefficient at $(u,v)$ can be computed as:
	\begin{equation*}\small
		\setlength{\abovedisplayskip}{3pt}
		\setlength{\belowdisplayskip}{3pt}
		\label{eqn:dft}
		\begin{split}
			\mathcal{F}_x(u,v) & = \sum\nolimits_{n=1}^{N}\sum\nolimits_{m=1}^{N}x(n,m)\cdot e^{i\theta}, \\
			& = \sum\nolimits_{n=1}^{N}\sum\nolimits_{m=1}^{N}x(n,m)\cdot (\cos{\theta} + i \cdot \sin{\theta}),
		\end{split}
	\end{equation*}
	where $\theta = -2\pi(un+vm)/N$. Then, the real and the imaginary parts of $\mathcal{F}_x(u,v)$ can be rewritten as:
	\begin{equation*}\small
		\setlength{\abovedisplayskip}{3pt}
		\setlength{\belowdisplayskip}{3pt}
		\label{eqn:temp}
		\begin{split}
			\mathcal{R}_x(u,v) & = \sum_{\cos{\theta} \ge 0} \cos{\theta} \cdot x(n,m) + \sum_{\cos{\theta} < 0} \cos{\theta} \cdot x(n,m), \\
			\mathcal{I}_x(u,v) & = \sum_{\sin{\theta} \ge 0} \sin{\theta} \cdot x(n,m) + \sum_{\sin{\theta} < 0} \sin{\theta} \cdot x(n,m).
		\end{split}
	\end{equation*}

	\begin{figure}[!tb]
		\centering
		\setlength{\abovecaptionskip}{0.cm}
		\includegraphics[width=0.9\linewidth]{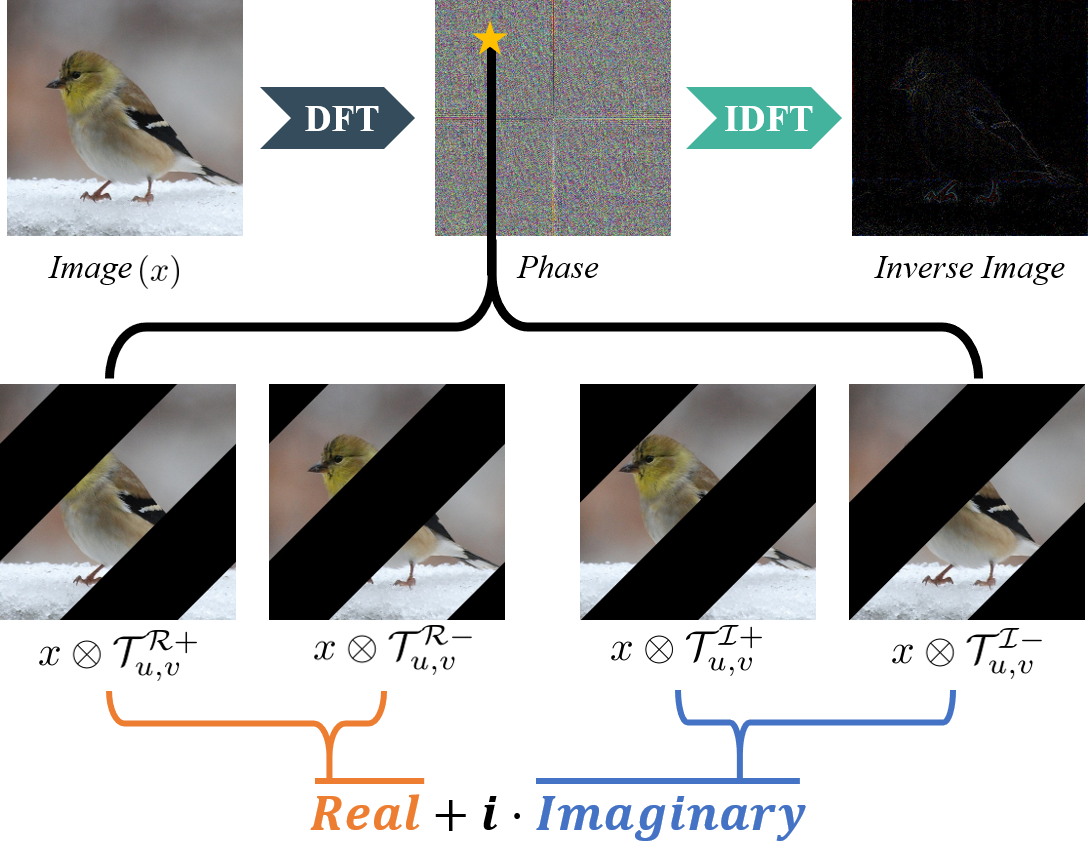}
		\caption{The four template-based contrasts for $(u,v)$ in Eq.\eqref{eqn:four}. Each Fourier coefficient is computed by dividing an image into two pairs of regions by the signs of real-part and the imaginary-part. These signs are encoded in spectral phase.}
		\label{fig:phase}
		\vspace{-4mm}
	\end{figure}
	
	The frequency in $(u,v)$ by Fourier transform can be interpreted as computing by four template-based contrasts:
	\begin{equation}\small
		\setlength{\abovedisplayskip}{3pt}
		\setlength{\belowdisplayskip}{3pt}
		\label{eqn:four}
		\begin{split}
			\mathcal{T}^\mathcal{R+}_{u,v}(x) & = \max (\cos{\theta},0), \mathcal{T}^\mathcal{R-}_{u,v}(x) = \max (-\cos{\theta},0), \\
			\mathcal{T}^\mathcal{I+}_{u,v}(x) & = \max (\sin{\theta},0), \mathcal{T}^\mathcal{I-}_{u,v}(x) = \max (-\sin{\theta},0). 
		\end{split}
	\end{equation}
	Moreover, we can define $4 \times N \times N$ templates for an image $x$ based on the signs of the real-part and the imaginary-part.
	A template-based example is shown in Figure \ref{fig:phase}.
	More examples for templates are shown in Appendix.
	
	Meanwhile, the phase spectrum $\mathcal{P}_x(u,v)$ for the image $x$ is equal to $\arctan (\frac{\mathcal{I}_x(u,v)}{\mathcal{R}_x(u,v)})$, which can be reinterpreted as:
	\begin{equation}\small
		\setlength{\abovedisplayskip}{3pt}
		\setlength{\belowdisplayskip}{3pt}
		\label{eqn:phase}
		\mathcal{P}_x(u,v) = \arctan (\frac{\sum x \otimes \mathcal{T}^{\mathcal{I+}}_{u,v} - \sum x \otimes \mathcal{T}^{\mathcal{I-}}_{u,v}}{\sum x \otimes \mathcal{T}^{\mathcal{R+}}_{u,v} - \sum x \otimes \mathcal{T}^{\mathcal{R-}}_{u,v}}).
	\end{equation}
	In Eq.\eqref{eqn:phase}, first, we can observe that the above four templates are encoded in the spectral phase. 
	Hence, all $4 \times N \times N$ templates are contained in the phase spectrum.
	This template-based contrast can help to explain the importance of the phase spectrum.
	Once the templates containing more targets without distractors are correctly estimated, the model can highly effectively locate the target objects \cite{li2015finding}.
	On the other hand, these templates in the phase spectrum could help to recover the structural information of the original image even without the original amplitude spectrum as shown in Figure \ref{fig:examples}.
	The robustness human visual system can also rely on this visible structured information for recognition.
	% Any coherent processing requires extreme care in order to maintain correct phase information \cite{ghiglia1998two}.
	% It is because of these templates in the phase that the phase contains the structural information of the picture
	% Secondly, the amplitude information $|\mathcal{R}_x|$ could be reasoned out by the phase spectrum once the imaginary-part is correctly estimated.
	% That is the reason that the most important features of a signal are preserved if only the phase spectrum is retained \cite{oppenheim1981importance}.

	\begin{figure*}[!tb]
		\centering
		\setlength{\abovecaptionskip}{0.cm}
		\includegraphics[width=0.9\linewidth]{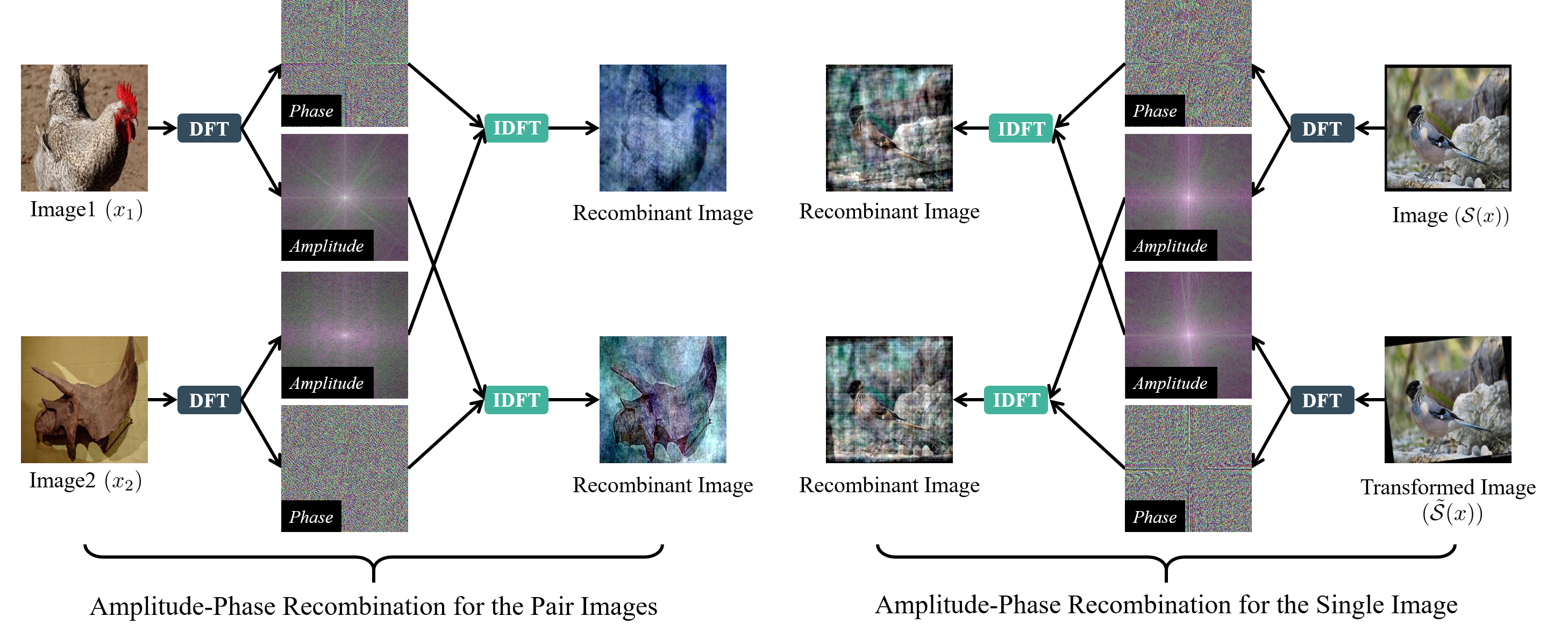}
		\caption{The two ways of Amplitude-Phase Recombination: APR-P and APR-S. The inversed images by APR-S are less different from the original image, compared with samples through ARP-P.}
		\label{fig:apr}
		\vspace{-1mm}
	\end{figure*}
	
	\section{Amplitude-Phase Recombination}
	\label{sec:method}
	% The quantitative experiments in Section \ref{sec:exp} also demonstrate that CNN can capture some effective phase information, which is also confirmed in \cite{liu2021spatial} with precise mathematical derivation.
	Motivated by the powerful generalizability of the human, we argue that reducing the dependence on the amplitude spectrum and enhancing the ability to capture phase spectrum can improve the robustness of CNN.
	% Effective training constraints can improve the ability of CNN to capture phase information, which is also confirmed in \cite{liu2021spatial} with precise mathematical derivation. 
	% keeping the phase information and increasing the amplitude variation can improve the robustness of the neural network..
	Therefore, we introduce a none-parameter data augmentation routine, termed as Amplitude-Phase Recombination (APR), constructing more effective training examples based on the single sample or pair samples.
	
	\begin{table*}[!tb]
		\caption{The adaptability test of common corruptions and surface variations. Average classification error as percentages. All values are percentages and the best results are indicated in bold.}
		\centering
		\small
		\label{tab:corruptions}
		\setlength{\tabcolsep}{0.8mm}{
			\begin{tabular}{ll|cccccc|cccc}
				\hline
				& & Standard & Cutout & Mixup & CutMix & Adv Training & APR-P & AutoAugment & AugMix & APR-S & ARP-SP \\
				\hline
				\hline
				\multirow{4}*{CIFAR-10-C} & AllConvNet & 30.8 & 32.9 & 24.6 & 31.3 & 28.1 & \textbf{21.5} & 29.2 & 15.0 & 14.8 & \textbf{11.5} \\
				& DenseNet & 30.7 & 32.1 & 24.6 & 33.5 & 27.6 & \textbf{20.3} & 26.6 & 12.7 & 12.3 & \textbf{10.3}  \\
				& WideResNet & 26.9 & 26.8 & 22.3 & 27.1 & 26.2 & \textbf{18.3} & 23.9 & 11.2 & 10.6 & \textbf{9.1} \\
				& ResNeXt & 27.5 & 28.9 & 22.6 & 29.5 & 27 & \textbf{18.5} & 24.2 & 10.9 & 11.0 & \textbf{9.1} \\
				\hline
				\multicolumn{2}{c|}{Mean} & 29.0 & 30.2 & 23.5 & 30.3 & 27.2 & \textbf{19.7} & 26 & 12.5 & 12.2 & \textbf{10.0} \\
				\hline
				\hline
				\multirow{4}*{CIFAR-100-C} & AllConvNet & 56.4 & 56.8 & 53.4 & 56.0 & 56.0 & \textbf{47.5} & 55.1 & 42.7 & 39.8 & \textbf{35.9} \\
				& DenseNet & 59.3 & 59.6 & 55.4 & 59.2 & 55.2 & \textbf{49.8} & 53.9 & 39.6 & 38.3 & \textbf{35.8} \\
				& WideResNet & 53.3 & 53.5 & 50.4 & 52.9 & 55.1 & \textbf{44.7} & 49.6 & 35.9 & 35.5 & \textbf{32.9} \\
				& ResNeXt & 53.4 & 54.6 & 51.4 & 54.1 & 54.4 & \textbf{44.2} & 51.3 & 34.9 & 33.7 & \textbf{31.0} \\
				\hline
				\multicolumn{2}{c|}{Mean} & 55.6 & 56.1 & 52.6 & 55.5 & 55.2 & \textbf{46.6} & 52.5 & 38.3 & 36.8 & \textbf{33.9} \\
				\hline
				\hline
		\end{tabular}}
		\vspace{-4mm}
	\end{table*}
	
	\textbf{APR for the Pair Samples (APR-P).}
	Firstly, $(x_i,y_i)$ and $(x_j, y_j)$ are two examples drawn at random from our training data.
	The main principle of APR is to change the amplitude spectrum as much as possible while keeping the phase spectrum and the corresponding labels unchanged.
	Hence, the APR-P could be defined as:
	\begin{equation}\small
		\setlength{\abovedisplayskip}{3pt}
		\setlength{\belowdisplayskip}{3pt}
		APR_P(x_i,x_j) = iDFT(\mathcal{A}_{x_j} \otimes e^{i \cdot \mathcal{P}_{x_i}}). 
	\end{equation}
	Then, the inversed training pair samples {\small $(APR_P(x_i,x_j),y_i)$} and {\small $(APR_P(x_j,x_i),y_j)$} are generated.
	Note that we use labels of phase as targets to allow the model to find the effective structured information in the phase spectrum.
	Meanwhile, through a variety of spectrum changes, the model gradually ignores the information from the imperceptible amplitude spectrum.
	It can be implemented by the way as Mixup \cite{guo2019mixup} that uses a single data loader to obtain one minibatch, and then APR-P is applied to the original minibatch and the minibatch after random shuffling.

	\textbf{APR for the Single Sample (APR-S)}.
	For a single training sample, we consider a set $\mathcal{S}$ consisting of $K$ different (random or deterministic) transformations, denoted $\mathcal{S} = \{ S_1, S_2, \dots S_K \}$. 
	Here, we attempt to consider that the sample $(x, y)$ and its transformed sample $\hat{x}$ are two different samples with the same label.
	The process of APR-S could be denoted as:
	\begin{equation}\small
		\setlength{\abovedisplayskip}{3pt}
		\setlength{\belowdisplayskip}{3pt}
		APR_S(\mathcal{S}(x_i), \tilde{\mathcal{S}}(x_i)) = iDFT(\mathcal{A}_{\tilde{\mathcal{S}}(x_i)} \otimes e^{i \cdot \mathcal{P}_{\mathcal{S}(x_i)}}), 
	\end{equation}
	where $\tilde{\mathcal{S}}$ and $\mathcal{S}$ are transformations set based on different random seeds or sequences.
	% Most transformation methods can change the information of amplitude and phase, but some of them, such as Perm \cite{tack2020csi}, could not change the amplitude information of the image.
	% These transformations should be avoided in the transformations set $\mathcal{S}$.
	
	Moreover, these two ways of amplitude-phase recombination could be used in combination and generate different gains for different data.
	Several examples from APR-P and APR-S are shown in Figure \ref{fig:apr}.

	\section{Experiments}
	\label{sec:exp}
	
	\textbf{Datasets.}
	CIFAR-10 and CIFAR-100 \cite{krizhevsky2009learning} datasets contain small $32$x$32$x$3$ color natural images, both with 50,000 training images and 10,000 testing images. CIFAR-10 has 10 categories, and CIFAR-100 has 100. The larger and more difficult ImageNet \cite{deng2009imagenet} dataset contains 1,000 classes of approximately 1.2 milion large-scale color images.
	
	In order to measure a model's resilience to common corruptions and surface variations, we evaluate methods on the CIFAR-10-C, CIFAR-100-C, and ImageNet-C datasets \cite{hendrycks2019benchmarking}. These datasets are constructed by corrupting the original CIFAR and ImageNet testsets. For each dataset, there are a total of 15 \emph{noise}, \emph{blur}, \emph{weather}, and \emph{digital} corruption types, each appearing at 5 severity levels or intensities. Since these datasets are used to measure network behavior under data shift, these 15 corruptions are not introduced into the training procedure.
	
	To measure the ability for OOD detection, we consider CIFAR-10 as in-distribution and the following datasets as OOD: SVHN \cite{netzer2011reading}, resized LSUN and ImageNet \cite{liang2018enhancing}, CIFAR-100 \cite{krizhevsky2009learning}. 
	% Moreover, ImageNet-O \cite{hendrycks2019natural} is adopted as the out-of-distribution dataset for ImageNet-1K. ImageNet-O includes 2K examples from ImageNet-22K \cite{ILSVRC15} excluding ImageNet-1K.
	
	\begin{table*}[!tb]
		\caption{The experiment of distinguishing in- and various OOD data for image classification. The best results are indicated in bold.}
		\centering
		\small
		\label{tab:ood}
		\begin{tabular}{l|c|cccccc|c}
			\hline
			\multirow{2}*{Method} & \multirow{2}*{Test acc.} & \multicolumn{6}{c|}{CIFAR-10 $\longrightarrow$} & \multirow{2}*{Mean} \\
			& & SVHN & LSUN & ImageNet & LSUN(FIX) & ImageNet(FIX) & CIFAR100 &  \\
			\hline
			\hline
			Cross Entropy (CE) & 93.0 & 88.6 & 90.7 & 88.3 & 87.5 & 87.4 & 85.8 & 88.1 \\
			CE w/ Cutout \cite{devries2017improved} & 95.8 & 93.6 & \textbf{94.5} & 90.2 & \textbf{92.2} & 89.0 & 86.4 & 91.0 \\
			CE w/ Mixup \cite{guo2019mixup} & \textbf{96.1} & 78.1 & 80.7 & 76.5 & 80.7 & 76.0 & 74.9 & 77.8 \\
			CE w/ APR-P & 95.0 & \textbf{98.1} & 93.7 & \textbf{95.2} & 91.4 & \textbf{91.1} & \textbf{88.9} & \textbf{93.1} \\
			\hline
			SupCLR \cite{khosla2020supervised} & 93.8 & 97.3 & 92.8 & 91.4 & 91.6 & 90.5 & 88.6 & 92.0 \\
			CSI \cite{tack2020csi} & 94.8 & 96.5 & 96.3 & 96.2 & 92.1 & 92.4 & \textbf{90.5} & 94.0 \\
			CE w/ APR-S & 95.1 & 90.4 & 96.1 & 94.2 & 90.9 & 89.1 & 86.8 & 91.3 \\
			CE w/ APR-SP & \textbf{95.6} & \textbf{97.7} & \textbf{97.9} & \textbf{96.3} & \textbf{93.7} & \textbf{92.8} & 89.5 & \textbf{94.7} \\
			\hline
			\hline
		\end{tabular}
		\vspace{-4mm}
	\end{table*}
	
	\subsection{CIFAR-10 and CIFAR-100}
	
	% \subsubsection{Training Setup}
	% \noindent
	\textbf{Training Setup.}
	For a model's resilience to common corruptions and surface variations, we adopt various architectures including an All Convolutional Network \cite{salimans2016weight}, a DenseNet-BC ($k=2,d=100$) \cite{huang2017densely}, a 40-2 Wide ResNet \cite{zagoruyko2016wide}, and a ResNeXt-29 (32x4) \cite{xie2017aggregated}. All networks use an initial learning rate of 0.1 which decay every 60 epochs. All models require 200 epochs for convergence. We optimize with stochastic gradient descent using Nesterov momentum \cite{sutskever2013importance}. All input images are processed with "Standard" random left-right flipping and cropping prior to any augmentations. For the data augmentations of APR-S, we adopt those used in \cite{hendrycks2019augmix} which is shown in Appendix. 
	For the OOD detection, we use ResNet-18 \cite{he2016deep} with the same training strategies above. 
	The data augmentations are set up the same as the above. 
	We report the Area Under the Receiver Operating Characteristic curve (AUROC) \cite{hendrycks17baseline} as a threshold-free evaluation metric for a detection score.
	We divide all methods into two categories, one is to add one augmentation on the basis of standard augmentations (random left-right flipping, and cropping), and the other is to add a combination of multiple augmentations as \cite{cubuk2018autoaugment,hendrycks2019augmix}.
	
	% \subsubsection{Results}
	\textbf{Common Corruptions and Surface Variations.}
	We first evaluate all methods with common corruptions and surface variations, such as noise, blur, weather, and digital.
	Compared to the Mixup or CutMix based on pair images, our APR-P with exchanging amplitude spectrum in pair images achieves 6\% lower absolute corruption error for CIFAR-100 as shown in Table \ref{tab:corruptions}.
	For methods based on a combination of multiple augmentations, our APR-S of the single image with just Cross-Entropy loss (CE) performs better than AugMix with simply mixing random augmentations and using the Jensen-Shannon loss substantially.
	When combining our method for single and pair images, the APR-SP achieves 5\% performance improvement compared with AugMix in CIFAR-100.
	In addition to surpassing numerous other data augmentation techniques, Table \ref{tab:corruptions} also demonstrates that these gains come from simple recombination of amplitude and phase without a complex mixup strategy.
	More comparisons and results about test accuracy are shown in Appendix.

	\textbf{Out-of-Distribution Detection.}
	We compare APR with those augmentations (Cutout, and Mixup) and those several training methods, the cross-entropy, supervised contrastive learning (SupCLR) \cite{khosla2020supervised}, and state-of-the-art method contrasting shifted instances (CSI) \cite{tack2020csi}. 
	Since our goal is to calibrate the confidence, the maximum softmax probability is used to detect OOD samples. 
	Table \ref{tab:ood} shows the results. 
	Firstly, APR-P consistently improves 2\% AUROC  than Cutout on CIFAR-10 while maintaining test accuracy. 
	Then, after combining APR based on single and pair images, APR-SP exceeds CSI and gains in almost all OOD tasks.
	APR promotes CNN to pay more attention to the phase spectrum so that some OOD samples that affect CNN's decision-making in amplitude spectrum could be detected effectively.
	
	\begin{table}[htb]
		\caption{Performance of ResNet-18 against with AutoAttack \cite{croce2020reliable}. }
		\centering
		\small
		% \footnotesize
		\label{tab:attack}
		\begin{tabular}{l|c|c}
			\hline
			\multirow{2}*{Method} & \multirow{2}*{Clean} & AutoAttack\cite{croce2020reliable} \\
			& & $l_{\inf} (\epsilon = 8/255)$  \\
			\hline\hline
			FSGM \cite{wong2019fast} & 83.3 & 43.2  \\
			FSGM w/ Cutout & 81.3 & 41.6  \\
			% FSGM w/ Mixup & 78.1 & 0.0 \footnote{The Mixup is not suitable for the adversarial training of FSGM.} \\
			\hline
			FSGM w/ APR-P & \textbf{85.3} & 44.1  \\
			FSGM w/ APR-S & 83.5 & 45.0 \\
			FSGM w/ APR-SP & 84.3 & \textbf{45.7}  \\
			\hline
			\hline
		\end{tabular}
		\vspace{-4mm}
	\end{table}
	
	\textbf{Adversarial Attack.}
	Moreover, the phenomenon of CNN focusing on amplitude spectrum leads to a question of whether APR can improve the adversarial robustness of models.
	Here, we evaluate several augmentations against one adversarial attack, AutoAttack \cite{croce2020reliable}.
	Table \ref{tab:attack} shows the AutoAttack \cite{croce2020reliable} performance by combining different methods with revisiting adversarial training method of FSGM \cite{wong2019fast} on CIFAR10.
	The cutout is not able to effectively against adversarial attacks compared with the baseline with revisiting adversarial training method of FSGM \cite{wong2019fast}.
	On the contrary, APR could effectively against AutoAttack while maintaining test accuracy.
	Compared with APR-P, APR-S for single images achieves more improvement on AutoAttack.
	Furthermore, the combination of these two strategies achieves better performance.
	It is evident that APR-SP improves the ability of the original model not only on clean images but also against adversarial attacks.
	
	\begin{table*}[!tb]
		\caption{Test Error, Corruption Error ($CE_c$), and $mCE_c$ values for various methods with ResNet-50 on ImageNet-C. All values are percentages and the best results are indicated in bold.}
		\centering
		% 	\scriptsize
		\footnotesize
		% \small
		\label{tab:imagenet}
		\setlength{\tabcolsep}{0.8mm}{
			\begin{tabular}{l|c|ccc|cccc|cccc|cccc|c}
				\hline
				% \toprule
				\multirow{2}*{Method} & \multirow{2}*{Test Err.} & \multicolumn{3}{c|}{Noise} & \multicolumn{4}{c|}{Blur} & \multicolumn{4}{c|}{Weather} & \multicolumn{4}{c|}{Digital} & \multirow{2}*{mCE} \\
				% \cline{3-17}
				& & Gauss & Shot & Impulse & Defocus & Glass & Motion & Zoom & Snow & Frost & Fog & Bright & Contrast & Elastic & Pixel & JPEG & \\
				\hline
				\hline
				Standard & 23.9 & 79 & 80 & 82 & 82 & 90 & 84 & 80 & 86 & 81 & 75 & 65 & 79 & 91 & 77 & 80 & 80.6 \\
				Patch Uniform & 24.5 & 67 & 68 & 70 & 74 & 83 & 81 & 77 & 80 & 74 & 75 & 62 & 77 & 84 & 71 & 71 & 74.3 \\
				AutoAugment(AA) & 22.8 & 69 & 68 & 72 & 77 & 83 & 80 & 81 & 79 & 75 & 64 & 56 & 70 & 88 & 57 & 71 & 72.7 \\
				Random AA & 23.6 & 70 & 71 & 72 & 80 & 86 & 82 & 81 & 81 & 77 & 72 & 61 & 75 & 88 & 73 & 72 & 76.1 \\
				MaxBlur pool & 23.0 & 73 & 74 & 76 & 74 & 86 & 78 & 77 & 77 & 72 & 63 & 56 & 68 & 86 & 71 & 71 & 73.4 \\
				SIN & 27.2 & 69 & 70 & 70 & 77 & 84 & 76 & 82 & 74 & 75 & 69 & 65 & 69 & 80 & 64 & 77 & 73.3 \\
				AugMix & 22.4 & 65 & 66 & 67 & 70 & \textbf{80} & \textbf{66} & \textbf{66} & 75 & 72 & 67 & 58 & 58 & \textbf{79} & \textbf{69} & \textbf{69} & 68.4 \\
				\hline
				APR-S & 24.5 & 61 & 64 & 60 & 73 & 87 & 72 & 81 & 72 & 67 & 62 & 56 & 70 & 83 & 79 & 71 & 70.5 \\
				APR-P & 24.4 & 64 & 68 & 68 & 70 & 89 & 69 & 81 & 69 & 69 & 55 & 57 & 58 & 85 & 66 & 72 & 69.3 \\
				% AugMix & 25.2 & 61 & 62 & 61 & 69 & 77 & 63 & 72 & 66 & 68 & 63 & 59 & 52 & 74 & 60 & 67 & 64.9 \\
				APR-SP & 24.4 & \textbf{55} & \textbf{61} & \textbf{54} & \textbf{68} & 84 & 68 & 80 & \textbf{62} & \textbf{62} & \textbf{49} & \textbf{53} & \textbf{57} & 83 & 70 & \textbf{69} & \textbf{65.0} \\
				\hline
				DeepAugment \cite{hendrycks2020many} & 26.3 & 49 & 49 & 48 & 62 & 74 & 68 & \textbf{79} & 68 & 64 & 64 & 57 & 63 & \textbf{78} & 50 & 73 & 63.1 \\
				DeepAugment+APR-SP & 26.4 & \textbf{44} & \textbf{45} & \textbf{41} & \textbf{57} & \textbf{70} & \textbf{60} & \textbf{79} & \textbf{56} & \textbf{56} & \textbf{50} & \textbf{54} & \textbf{54} & \textbf{78} & \textbf{47} & \textbf{71} & \textbf{57.5} \\
				\hline
				\hline
				% \bottomrule
		\end{tabular}}
		\vspace{-4mm}
	\end{table*}
	
	\subsection{ImageNet Classsification}
	\textbf{Training Setup.}
	ResNet-50 \cite{he2016deep} is trained with an initial learning rate of 0.1 which decay every 30 epochs. 
	It is optimized with stochastic gradient descent using momentum $0.9$ \cite{qian1999momentum}, and requires 100 epochs for convergence.
	All input images are pre-processed with standard random cropping horizontal mirroring.
	For the data augmentations of APR-S, we adopt those used in \cite{hendrycks2019augmix} without augmentations such as \emph{contrast, color, brightness, sharpness, and Cutout}, which may overlap with the corruptions of ImageNet-C. 
	Following \cite{hendrycks2019augmix}, we utilize the convention of normalizing the corruption error by the corruption err of AlexNet \cite{krizhevsky2012imagenet}.
	Corruption Error ($CE_c$) is computed as $CE_c=\sum_{s=1}^{5}E_{c,s} / \sum_{s=1}^{5}E_{c,s}^{AlexNet}$.
	The average of the 15 corruption errors is as the Mean Corruption Error (${mCE}_c$).
	
	\begin{figure}[!tb]
		\centering
		\setlength{\abovecaptionskip}{0.cm}
		\subfigure[Original]{
			\begin{minipage}[t]{0.23\linewidth}
				\centering
				\includegraphics[width=\linewidth]{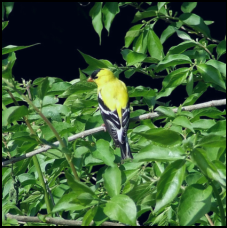}
				\label{fig:b1}
			\end{minipage}%
		}
		\subfigure[Fog]{
			\begin{minipage}[t]{0.23\linewidth}
				\centering
				\includegraphics[width=\linewidth]{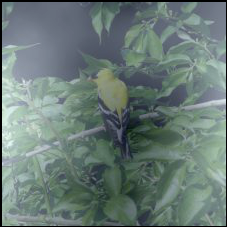}
				\label{fig:o2}
			\end{minipage}%
		}
		\subfigure[Standard]{
			\begin{minipage}[t]{0.23\linewidth}
				\centering
				\includegraphics[width=\linewidth]{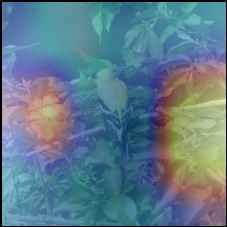}
				\label{fig:o3}
			\end{minipage}%
		}
		\subfigure[APR-SP]{
			\begin{minipage}[t]{0.23\linewidth}
				\centering
				\includegraphics[width=\linewidth]{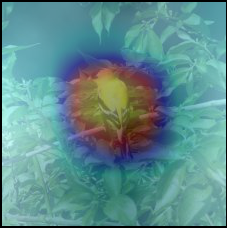}
				\label{fig:o4}
			\end{minipage}%
		}
		\caption{The Gradient-weighted Class Activation Mapping \cite{selvaraju2017grad} of the baseline and the proposed APR-SP for images with frog noise. Best viewed in color. APR-SP still is able to focus on the parts of the target object even in a heavy fog.}
		\label{fig:cam}
		\vspace{-4mm}
	\end{figure}
	
	\textbf{Results.}
	Our method APR-SP achieves 15\% improvement than the baseline $80.6\%$ ${mCE}_c$ while maintaining test accuracy. 
	Other methods such as AutoAugment and AugMix require a more complex combination strategy, while ours does not.
	Meanwhile, APR improves corruption robustness \cite{hendrycks2019augmix} and uncertainty estimates across almost every individual corruption and severity level while the performance of \emph{zoom blur} is comparable with most methods.
	APR-SP gets about 5\% improvement than APR-S and APR-P, and APR-SP with DeepAugment improves 6\% than the reproduced DeepAugment \cite{hendrycks2020many}.
	As shown in Figure \ref{fig:cam}, the CNN trained with APR-SP is able to focus on the parts of the target object for classification even in a heavy fog.
	These results demonstrate that scaling up APR from CIFAR to ImageNet also leads to state-of-the-art results in robustness and uncertainty estimation.
	
	\subsection{Labeled by Amplitude or Phase?}
	
\begin{figure}[!tb]
		\centering
		\setlength{\abovecaptionskip}{0.cm}
		\includegraphics[width=0.9\linewidth]{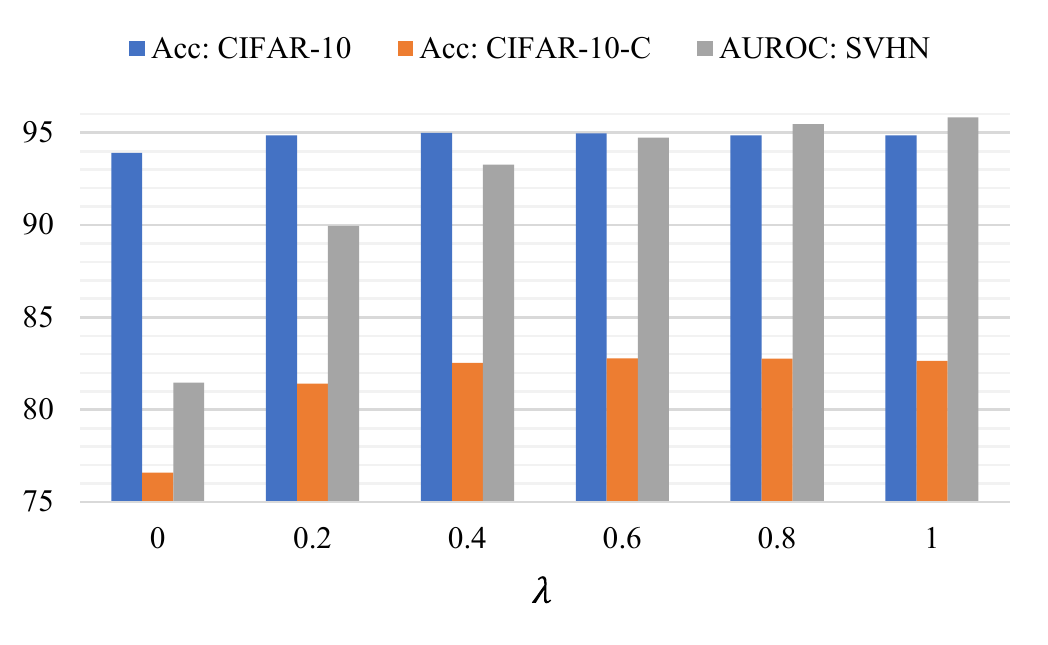}
		\caption{The performance of ResNet-18 for various distribution as different attention weights for the amplitude and phase spectrum.}
		\label{fig:labdma}
		\vspace{-4mm}
	\end{figure}

	For our proposed APR-P, we utilize the labels of phase spectrum in the pair samples.
	Naturally, we wish to explore the impact of using labels amplitude and phase separately.
	Here, we add a linear classifier layer in ResNet-18 to predict the labels of the amplitude spectrum.
	The model is trained for the sample $\bar{x}$ combined by the phase spectrum $\mathcal{P}_{x_i}$ and the amplitude spectrum $\mathcal{A}_{x_j}$ by optimizing:
	\begin{equation}\small
		\setlength{\abovedisplayskip}{3pt}
		\setlength{\belowdisplayskip}{3pt}
		\mathop{\arg\min}_{\theta}{ \ \lambda l(f_\mathcal{P}(\bar{x}; \theta), y_i) + (1-\lambda) \cdot l(f_\mathcal{A}(\bar{x}; \theta), y_j}).
	\end{equation}
	% where, $\lambda$ is set as $0.5$ in the training stage.
	Then, the final prediction is defined as $\bar{y}=\lambda f_\mathcal{P} + (1-\lambda)f_\mathcal{A}$.
	The recognition ability of the model to different distribution changes with $\lambda$ as shown in Figure \ref{fig:labdma}.
	With the enhancement of the weight of phase prediction, the accuracy of the model is improved, especially for common corruptions and surface variations, and OOD detection.
	Meanwhile, the detection ability of the model for OOD samples becomes stronger with the increase of phase attention.
	This result could further prove the correctness of our corollaries.
	
	\section{Conclusion \& Outlook}
	
	This paper proposes a series of quantitative and qualitative analyses to indicate that a robust CNN should be robust to the amplitude variance and pay more attention to the components related to the phase spectrum. Then,  a novel data augmentation method APR is proposed to force the CNN to pay more attention to the phase spectrum and achieves state-of-the-art performances on multiple generalizations and calibration tasks. Also,  a unified explanation is provided to the behaviors of both adversarial attack and the overconfidence of OOD by the CNN’s over-dependence on the amplitude spectrum.
	Looking forward, more research directions about phase could be exploited in the future era of computer vision research.
	One possible direction is to explore how to represent part-whole hierarchies \cite{hinton2021represent} in neural networks that rely on the phase spectrum.
	On the other hand, more CNN models \cite{sabour2017dynamic,roy2019towards} or convolution operations to capture more phase information are worth exploring.
	
	\noindent
	\textbf{Acknowledgments.} 
	% \section*{Acknowledgments}
	This work is partially supported by grants from the National Key R\&D Program of China under Grant 2020AAA0103501, and grants from the National Natural Science Foundation of China under contract No. 61825101 and No. 62088102.

	{\small
		\bibliographystyle{ieee_fullname}
		\bibliography{egbib}
	}
	
	\clearpage

	{\section*{\Large Appendix}}
	\setcounter{equation}{0}
	\setcounter{subsection}{0}
	\setcounter{section}{0}
	\renewcommand{\theequation}{A.\arabic{equation}}
	\renewcommand\thesection{\Alph{section}}

	%%%%%%%%% BODY TEXT
	% \appendix
	% \renewcommand{\appendixname}{Appendix~\Alph{section}}
	In the supplemental material, we firstly visualize the distributions of the corrupted samples, adversarial samples, and OOD samples in the frequency domain to validate the \textbf{Assumption} 2 in the main text in Section \ref{sec:study}. Then, several typical templates of the phase spectrum are shown in Section \ref{sec:templates_phase} to intuitionally explain the rationality of the proposed APR method, and the implementation details of the data augmentation of APR-S are listed in Section \ref{sec:augmentation}. 
	In Section \ref{sec:additional_results}, the Fourier analysis is provided to demonstrate the various gains from APR-P and APR-S, and the clean error analysis and OOD detection on ImageNet are also listed to clarify the excellent scalability of our method.
	
	\begin{figure}[!htb]
		\centering
		\subfigure[$\mathcal{A}^L$ of CS]{
			\begin{minipage}[t]{0.3\linewidth}
				\centering
				\includegraphics[width=\linewidth]{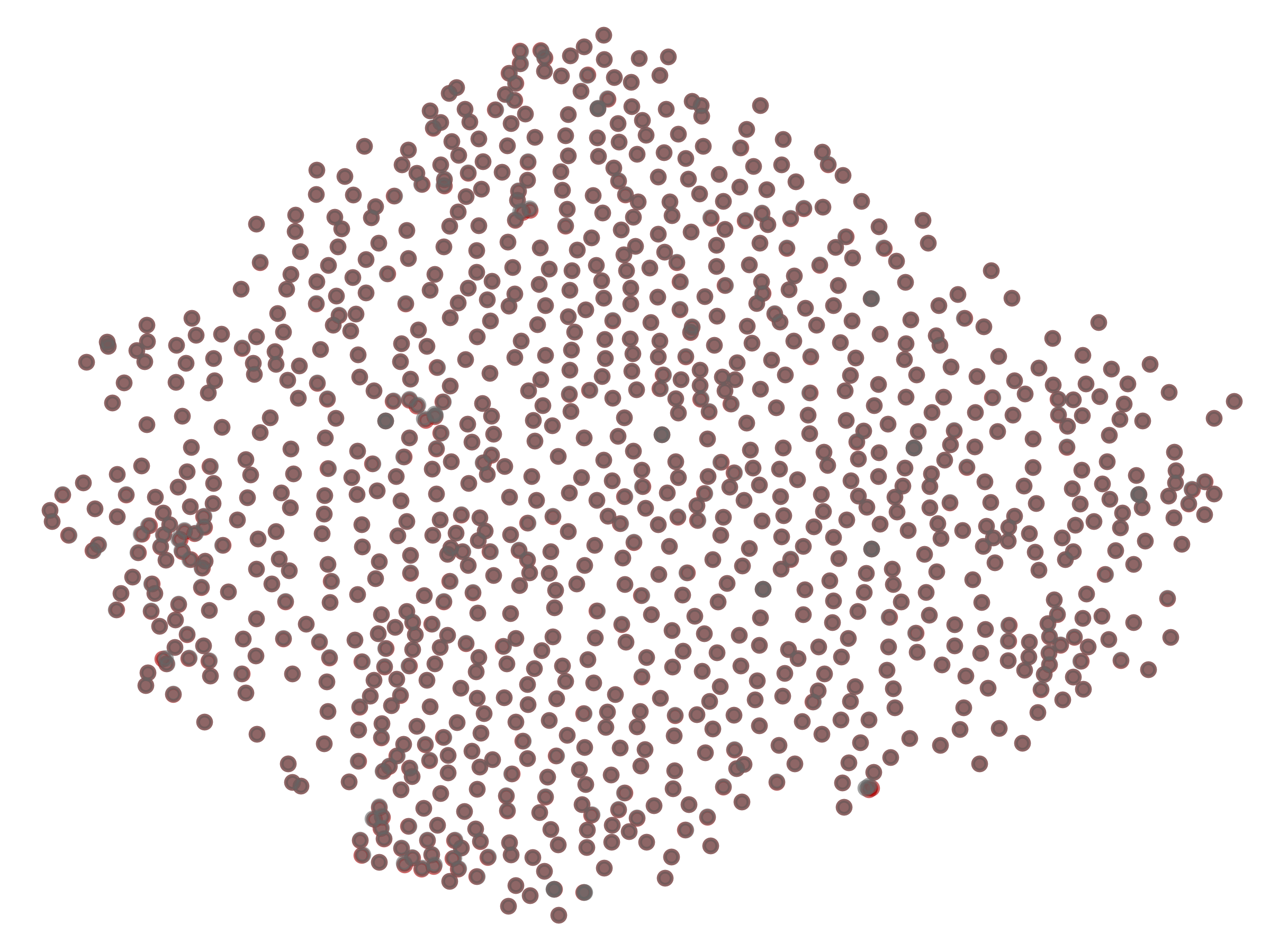}
				\label{fig:cs_low}
			\end{minipage}%
		}
		\subfigure[$\mathcal{A}^I$ of CS]{
			\begin{minipage}[t]{0.3\linewidth}
				\centering
				\includegraphics[width=\linewidth]{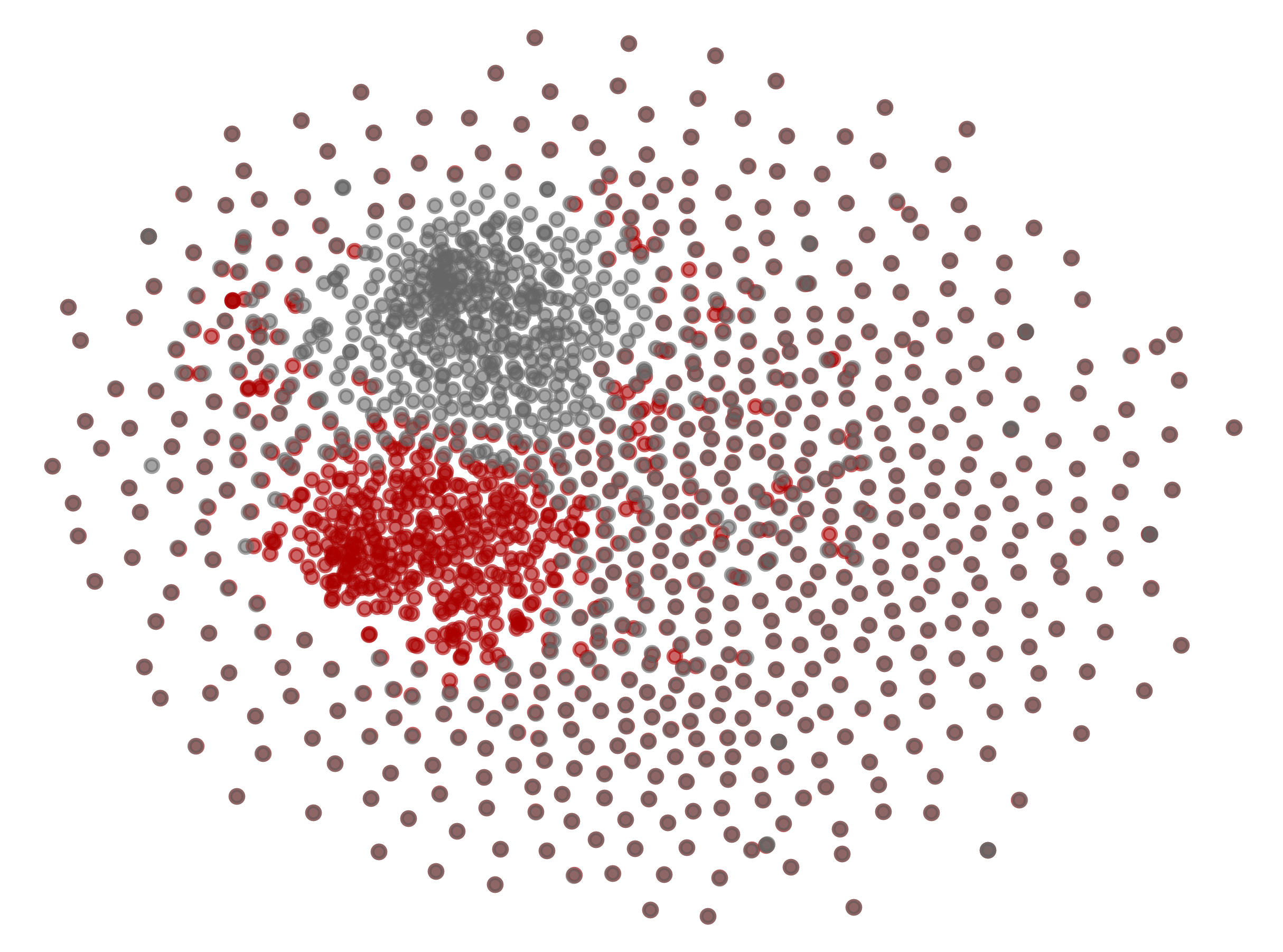}
				\label{fig:cs_mid}
			\end{minipage}%
		}
		\subfigure[$\mathcal{A}^H$ of CS]{
			\begin{minipage}[t]{0.3\linewidth}
				\centering
				\includegraphics[width=\linewidth]{figures/corruption/am_h.png}
				\label{fig:cs_hig}
			\end{minipage}%
		}
		\subfigure[$\mathcal{A}^L$ of AA]{
			\begin{minipage}[t]{0.3\linewidth}
				\centering
				\includegraphics[width=\linewidth]{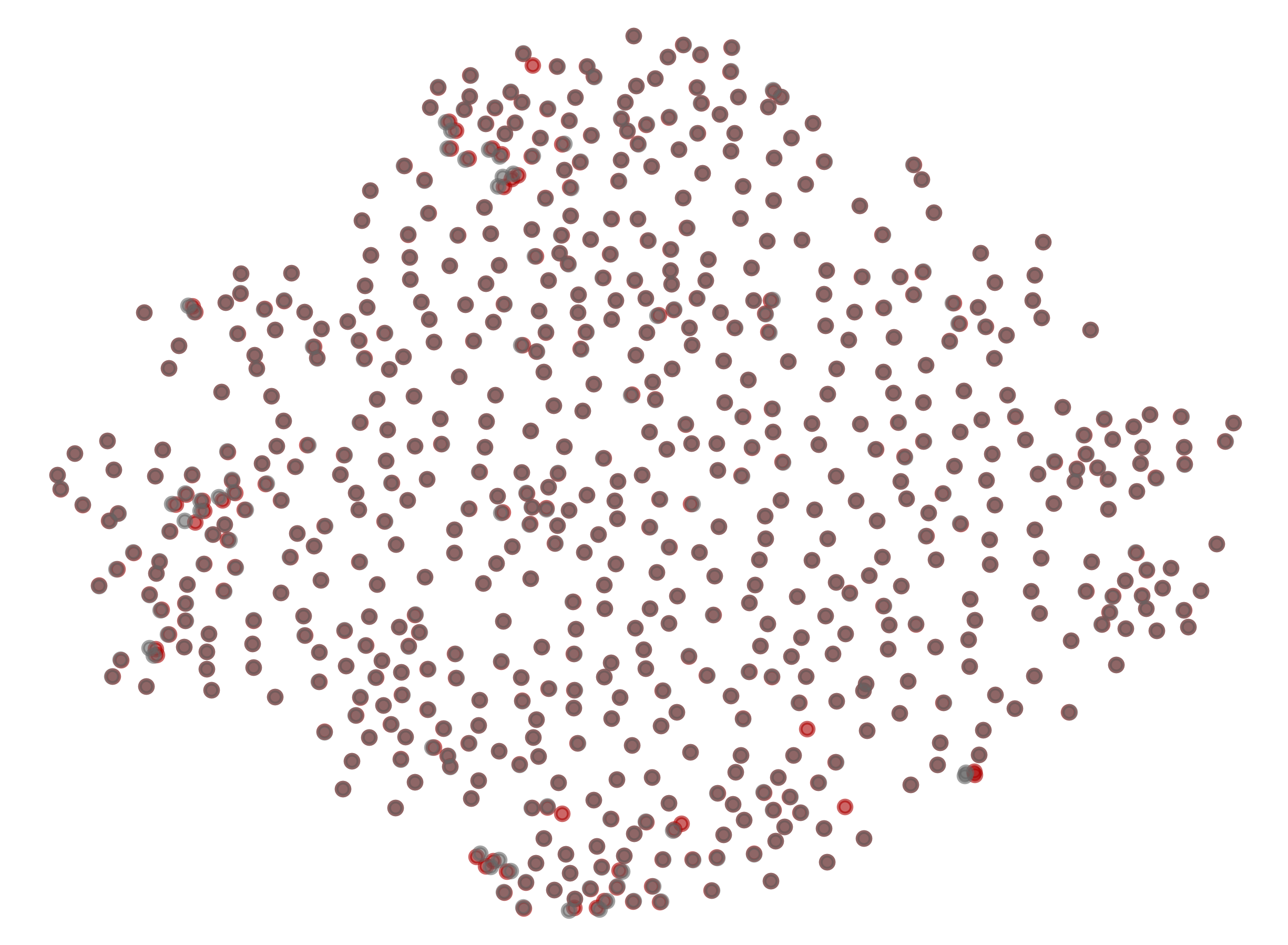}
				\label{fig:ad_low}
			\end{minipage}%
		}
		\subfigure[$\mathcal{A}^I$ of AA]{
			\begin{minipage}[t]{0.3\linewidth}
				\centering
				\includegraphics[width=\linewidth]{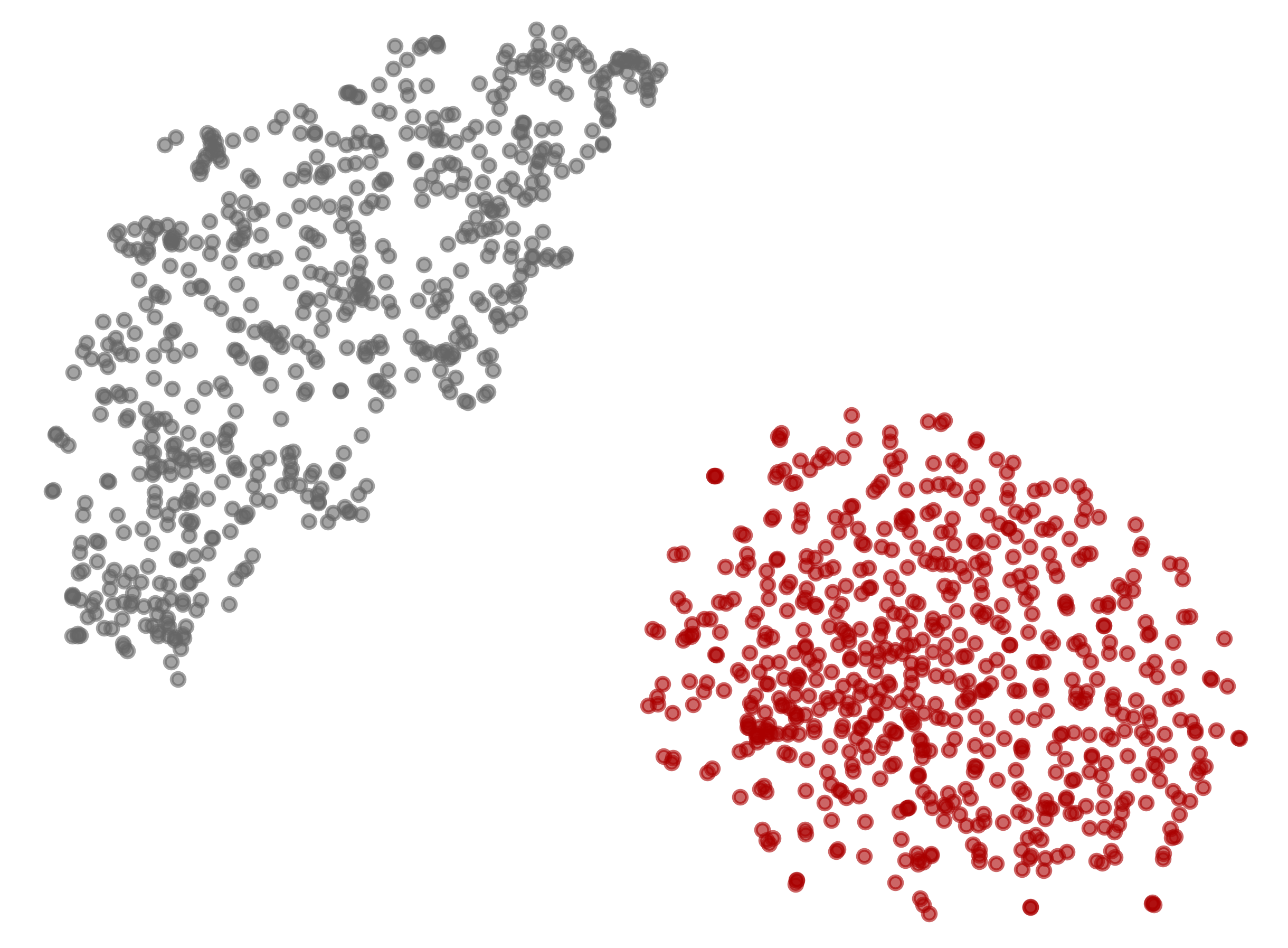}
				\label{fig:ad_mid}
			\end{minipage}%
		}
		\subfigure[$\mathcal{A}^H$ of AA]{
			\begin{minipage}[t]{0.3\linewidth}
				\centering
				\includegraphics[width=\linewidth]{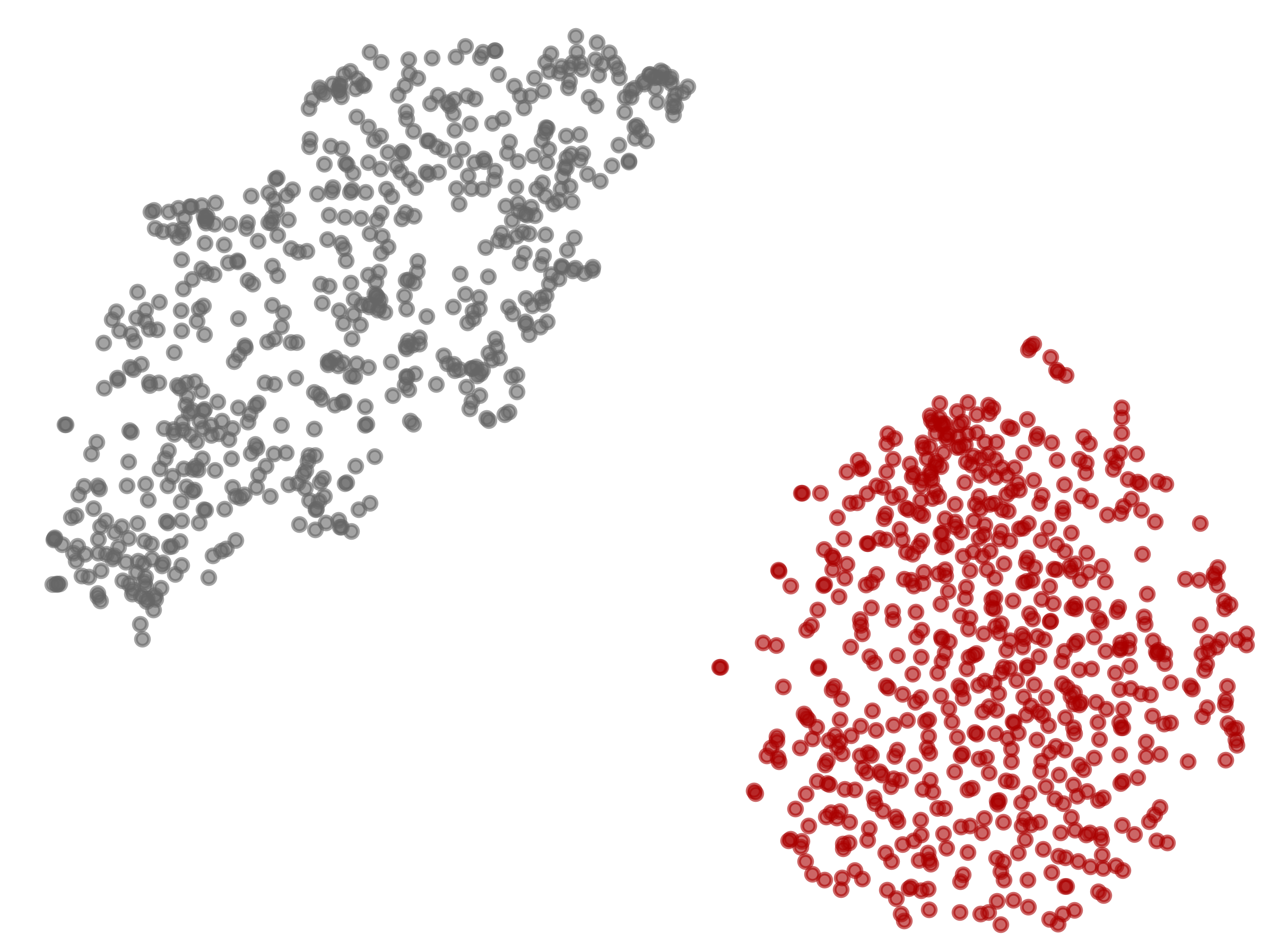}
				\label{fig:ad_hig}
			\end{minipage}%
		}
		\subfigure[$\mathcal{A}^L$ of OOD]{
			\begin{minipage}[t]{0.3\linewidth}
				\centering
				\includegraphics[width=\linewidth]{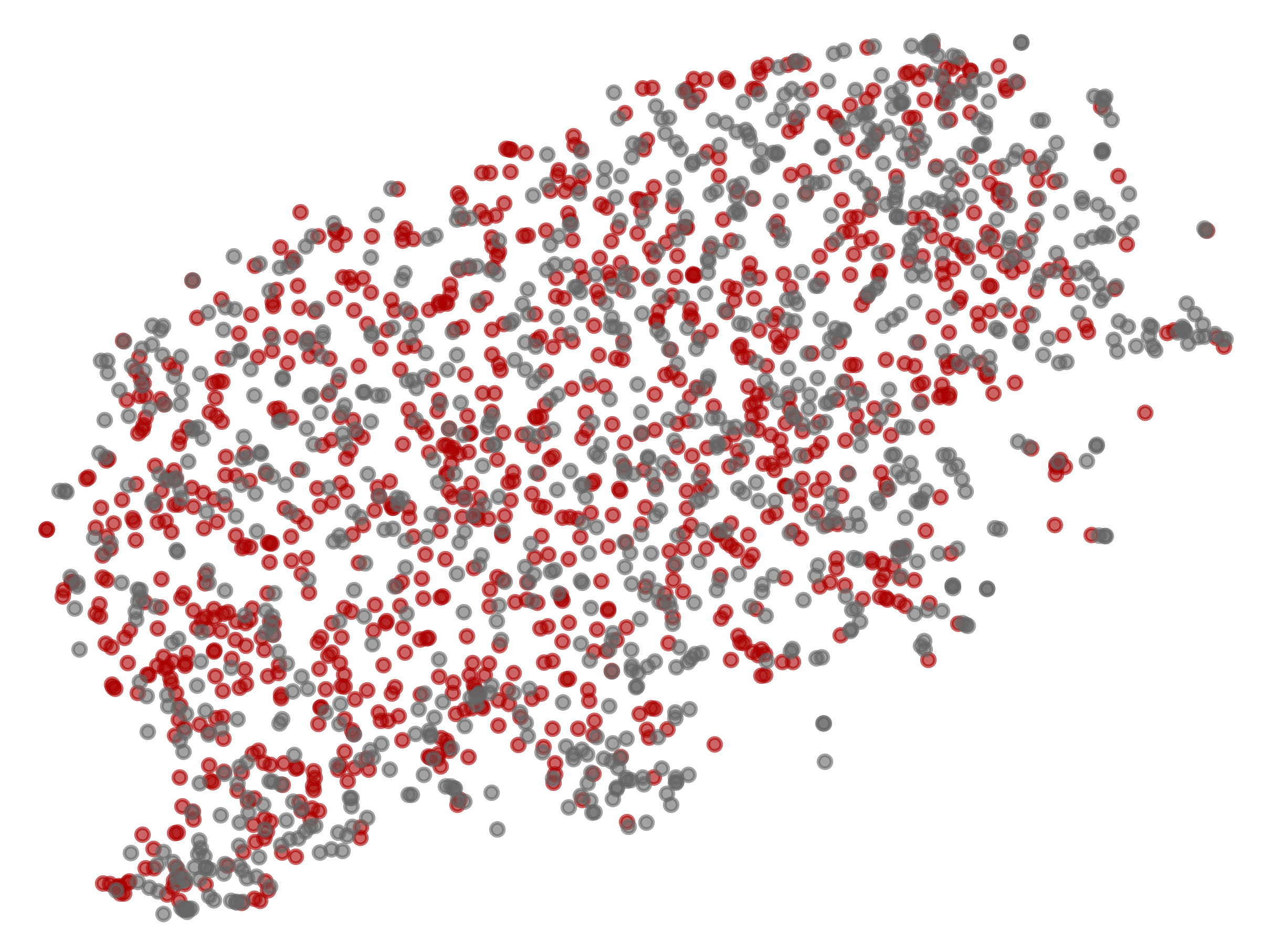}
				\label{fig:ood_low}
			\end{minipage}%
		}%
		\subfigure[$\mathcal{A}^I$ of OOD]{
			\begin{minipage}[t]{0.3\linewidth}
				\centering
				\includegraphics[width=\linewidth]{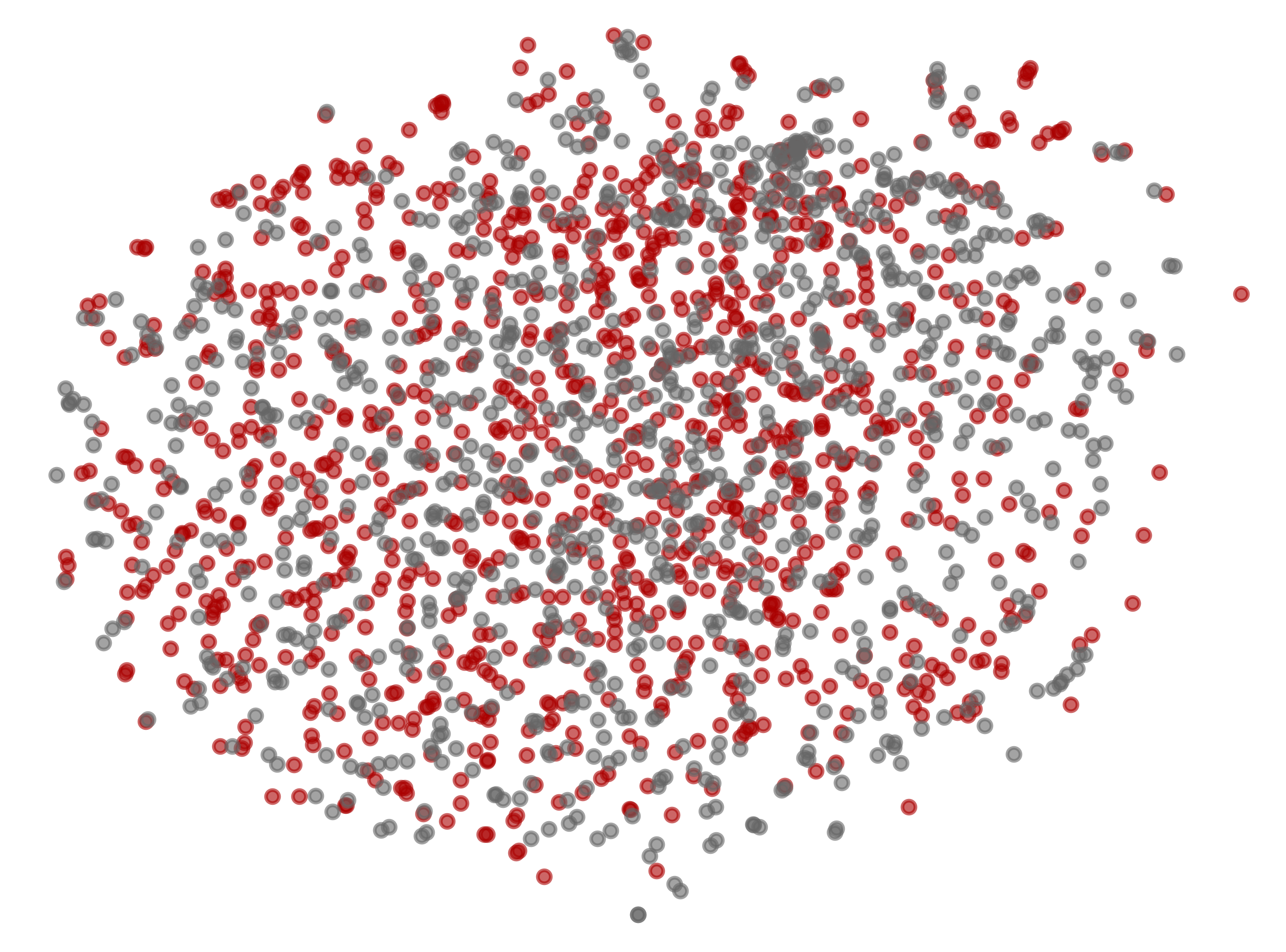}
				\label{fig:ood_mid}
			\end{minipage}%
		}%
		\subfigure[$\mathcal{A}^H$ of OOD]{
			\begin{minipage}[t]{0.3\linewidth}
				\centering
				\includegraphics[width=\linewidth]{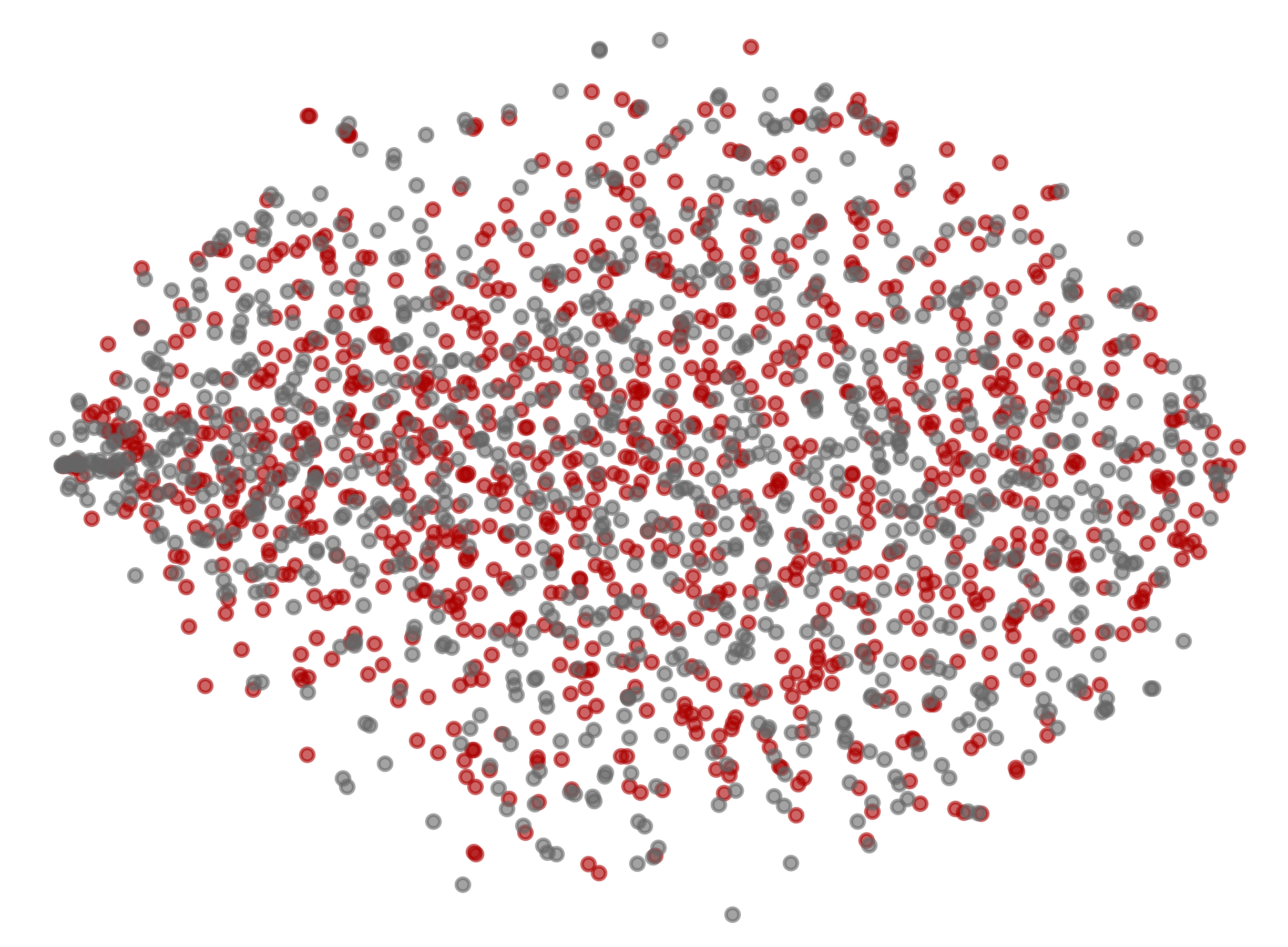}
				\label{fig:ood_hig}
			\end{minipage}%
		}%
		\caption{The T-SNE \cite{van2008visualizing} visualization of the different types of the amplitude spectrum. Red represents the original image or in-distribution (ID) samples in CIFAR-10, and gray represents Corrupted Samples (CS), the samples generated by Adversarial Attacks (AA) from CIFAR-10, or OOD samples from CIFAR-100.}
		\label{fig:examples2}
	\end{figure}
	
	\section{More Studies on the Frequency Domain.}
	\label{sec:study}
	
	Here, we show more studies of the different types of the amplitude spectrum.
	For all samples in CIFAR, we generate the low-frequency, intermediate-frequency and high-frequency counterparts with $r$ for non-zero parts set to $[0,8]$, $[8,16]$ and $[16,16\sqrt{2}]$, respectively.
	In Figure \ref{fig:examples2}, we show the amplitude spectrum distributions of low-frequency, intermediate-frequency, and high-frequency from original samples, their corrupted samples, adversarial samples, and OOD samples respectively.
	
	Firstly, for \emph{corrupted samples} and \emph{adversarial samples} from a single category, we could observe that their amplitude spectrum in high-frequency and intermediate-frequency has different distribution with the original samples even if only invisible noises are introduced.  
	Moreover, the amplitude spectrums in low-frequency of corrupted samples and adversarial samples are indistinguishable from the original images. It also explains that CNN captures the high-frequency image components for classification \cite{wang2020high}.
	Hence, CNN would make a wrong prediction for 'similar' images (corruption and adversarial samples) when the parts of the amplitude spectrum are changed.
	
	Then, for the \emph{OOD samples} from CIFAR-100, it is evident that any type of the amplitude spectrum of in-distribution and out-of-distribution could be not able to distinguish.
	CNN focusing on the amplitude spectrum has a huge risk that any OOD sample with a similar amplitude part of in-distribution samples would be classified as an in-distribution sample.
	Hence, CNN would be overconfident for some out-of-distributions when similar amplitude information appears.
	
	Overall, the above analyses explain our \textbf{Assumption} 2 in the main text, that the counter-intuitive behaviors of the sensitivity to common perturbations and the overconfidence of OOD maybe both be related to CNN's over-dependence on the amplitude spectrum.
	We do not focus on the high frequency only, because the OOD samples may come from the similarity of any amplitude part as shown in Figure \ref{fig:examples2}.
	As a result, the focusing of some parts of the amplitude spectrum may create an invisible way to attack CNN, such as the adversarial attack and various corruptions (Corollary 1), and the amplitude attack or OOD attack (Corollary 2).

	\section{Templates from the Phase Spectrum}
	\label{sec:templates_phase}
	
	\begin{figure}[!tb]
		\centering
		\subfigure[Positive Real Part $\mathcal{T}^\mathcal{R+}$]{
			\begin{minipage}[t]{0.45\linewidth}
				\centering
				\includegraphics[width=\linewidth]{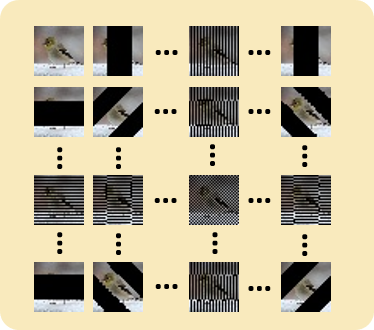}
				\label{fig:real_p}
			\end{minipage}%
		}
		\subfigure[Negative Real Part $\mathcal{T}^\mathcal{R-}$]{
			\begin{minipage}[t]{0.45\linewidth}
				\centering
				\includegraphics[width=\linewidth]{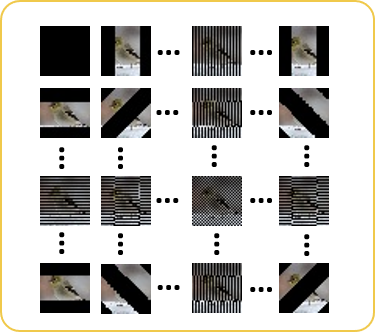}
				\label{fig:real_n}
			\end{minipage}%
		}
		\subfigure[Positive Imaginary Part $\mathcal{T}^\mathcal{I+}$]{
			\begin{minipage}[t]{0.45\linewidth}
				\centering
				\includegraphics[width=\linewidth]{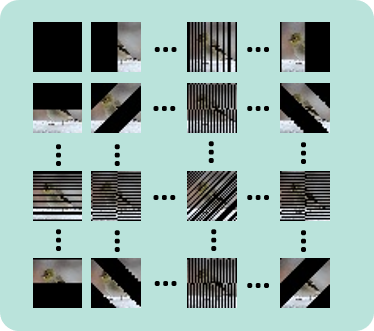}
				\label{fig:imag_p}
			\end{minipage}%
		}%
		\subfigure[Negative Imaginary Part $\mathcal{T}^\mathcal{I-}$]{
			\begin{minipage}[t]{0.45\linewidth}
				\centering
				\includegraphics[width=\linewidth]{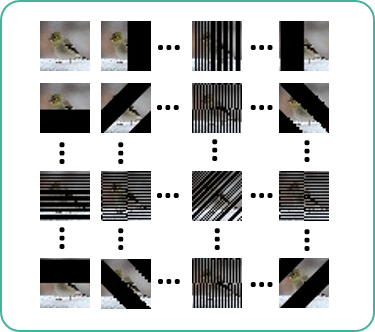}
				\label{fig:imag_n}
			\end{minipage}%
		}%
		\caption{Fourier transform can be interpreted as dividing an image with $4 \times N^2$ templates for contrast computation.}
		\label{fig:templates}
	\end{figure}
	
	To better reveal the role of the phase spectrum, we analyze the phase spectrum in the different frequency domains.
	When DFT is applied on a gray image, there are totally $4 \times N^2$ templates which are used to compute $2 \times N^2$ contrast scores (as shown in Figure \ref{fig:templates}).
	Consequently, the frequency spectrum stores contrast values obtained at multiple scales and directions.
	The classification or other visual tasks could benefit from capturing the difference between targets and distractors by these templates.
	The templates of the lowest frequencies divide images into large regions, which are "coarse" partitions.
	Then, the templates of the highest frequencies provide "fine" partitions that achieve only high responses to noises and textures.
	In addition, the templates of the intermediate frequencies provide "moderate" partitions which may include the target object, and it has also been proved that it's beneficial for fixation prediction in \cite{li2015finding}.
	These templates in the phase spectrum could help to recover the structural information of the original image even without the original amplitude spectrum \cite{oppenheim1981importance}.
	The robustness human visual system can also rely on this visible structured information for recognition \cite{oppenheim1981importance,liu2021spatial}.
	
	\section{Augmentation Operations}
	\label{sec:augmentation}
	
\begin{figure}[!tb]
		\centering
		\subfigure[AutoContrast]{
			\begin{minipage}[t]{0.3\linewidth}
				\centering
				\includegraphics[width=\linewidth]{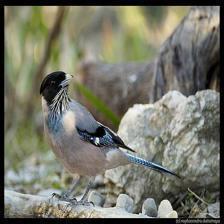}
				\label{fig:single_1}
			\end{minipage}%
		}
		\subfigure[Equalize]{
			\begin{minipage}[t]{0.3\linewidth}
				\centering
				\includegraphics[width=\linewidth]{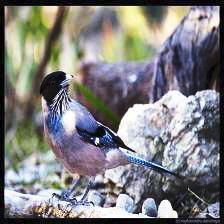}
				\label{fig:single_2}
			\end{minipage}%
		}
		\subfigure[Posterize]{
			\begin{minipage}[t]{0.3\linewidth}
				\centering
				\includegraphics[width=\linewidth]{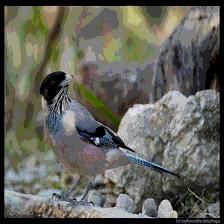}
				\label{fig:single_3}
			\end{minipage}%
		}\\
		\subfigure[Rotate]{
			\begin{minipage}[t]{0.3\linewidth}
				\centering
				\includegraphics[width=\linewidth]{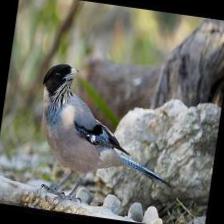}
				\label{fig:single_4}
			\end{minipage}%
		}%
		\subfigure[Solarize]{
			\begin{minipage}[t]{0.3\linewidth}
				\centering
				\includegraphics[width=\linewidth]{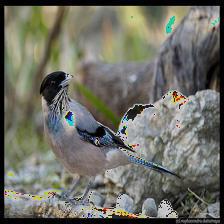}
				\label{fig:single_5}
			\end{minipage}%
		}%
		\subfigure[Shear $x$]{
			\begin{minipage}[t]{0.3\linewidth}
				\centering
				\includegraphics[width=\linewidth]{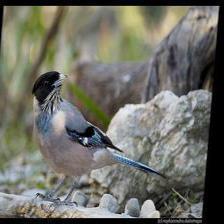}
				\label{fig:single_6}
			\end{minipage}%
		}\\
		\subfigure[Shear $y$]{
			\begin{minipage}[t]{0.3\linewidth}
				\centering
				\includegraphics[width=\linewidth]{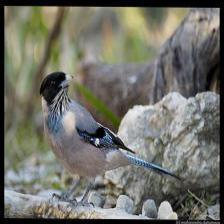}
				\label{fig:single_7}
			\end{minipage}%
		}%
		\subfigure[Translate $x$]{
			\begin{minipage}[t]{0.3\linewidth}
				\centering
				\includegraphics[width=\linewidth]{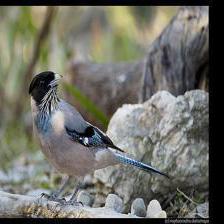}
				\label{fig:single_8}
			\end{minipage}%
		}%
		\subfigure[Translate $y$]{
			\begin{minipage}[t]{0.3\linewidth}
				\centering
				\includegraphics[width=\linewidth]{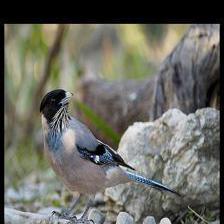}
				\label{fig:single_9}
			\end{minipage}%
		}%
		\caption{Illustration of augmentation operations applied to the same image.}
		\label{fig:aug}
	\end{figure}

	The augmentation operations used in APR-S are same with \cite{hendrycks2019augmix} as shown in Figure \ref{fig:aug}.
	We do not use \emph{contrast}, \emph{color}, \emph{brightness}, \emph{sharpness} and \emph{Cutout} as they may overlap with the corruptions of CIFAR-C and ImageNet-C.
	
	\section{Additional Results}
	\label{sec:additional_results}
	
	\begin{figure*}[!htb]
		\centering
		\includegraphics[width=\linewidth]{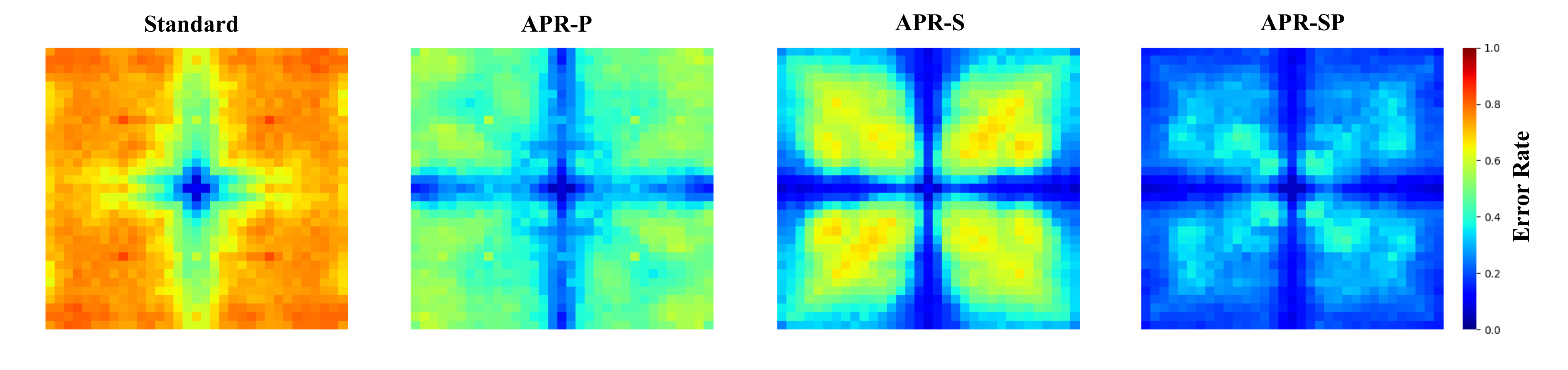}
		\caption{Model (Wide ResNet) sensitivity to the additive noise aligned with different Fourier basis vectors on CIFAR-10 validation images.
			We fix the additive noise to have $L_2$ norm $15$ and evaluate four methods: a standard trained model, APR-P, APR-S, APR-SP.
			Error rates are averaged over $1000$ randomly sampled images from the test set.
			The standard trained model is highly sensitive to the additive noise in all but the lowest frequencies.
			APR-SP could substantially improve robustness to most frequency perturbations.
		}
		\label{fig:fourier}
	\end{figure*}

	\begin{table*}[!thb]
		\caption{CIFAR-10 Clean Error. All values are percentages and the best results are indicated in bold.}
		\centering
		\small
		\label{tab:clean}
		\setlength{\tabcolsep}{0.8mm}{
			\begin{tabular}{ll|ccccccc|ccc}
				\hline
				& & Standard & Cutout & Mixup & CutMix & AutoAugment & Adv Training & AugMix & APR-P & APR-S & APR-SP \\
				\hline
				\hline
				\multirow{4}*{CIFAR-10-C} & AllConvNet & 6.1 & 6.1 & 6.3 & 6.4 & 6.6 & 18.9 & 6.5 & \textbf{5.5} & 6.5& 5.7 \\
				& DenseNet & 5.8 & \textbf{4.8} & 5.5 & 5.3 & \textbf{4.8} & 17.9 & 4.9 & 5.0 & 5.1 & \textbf{4.8} \\
				& WideResNet & 5.2 & 4.4 & 4.9 & 4.6 & 4.8 & 17.1 & 4.9 & 4.8 & 5.0 & \textbf{4.3} \\
				& ResNeXt & 4.3 & 4.4 & 4.2 & \textbf{3.9} & 3.8 & 15.4 & 4.2 & 4.5 & 4.5 & \textbf{3.9} \\
				\hline
				\multicolumn{2}{c|}{Mean} & 5.4 & 4.9 & 5.2 & 5.0 & 5.0 & 17.3 & 5.1 & 5.0 & 5.2 & \textbf{4.7} \\
				\hline
				\hline
		\end{tabular}}
	\end{table*}
	
	\subsection{Fourier Analysis} 
	
	In order to better understand the reliance of our methods on different frequencies, here we measure the model sensitivity to the additive noise at differing frequencies.
	We add a total of $33 \times 33$ Fourier basis vector to the CIFAR-10 test set, one at a time, and record the resulting error rate after adding each Fourier basis vector.
	Each point in the $33 \times 33$ sensitivity heatmap shows the error rate on the CIFAR-10 test set after it has been perturbed by a single Fourier basis vector.
	Points corresponding to the low-frequency vectors are shown in the center of the heatmap, whereas the high-frequency vectors are farther than the center.
	
	In Figure \ref{fig:fourier}, we observe that the standard model is robust to the low-frequency perturbations but severely lacks robustness to the high-frequency perturbations, where the error rates exceed $80\%$.
	Then, the model trained by APR-P is more robust to all frequencies, especially to the low and intermediate frequencies.
	Moreover, the model trained by APR-S maintains robustness to low-frequency perturbations and improves robustness to the high-frequency perturbations, but is still sensitive to the additive noise in the intermediate frequencies.
	This further explains the experiments in Section 5.1.2 (main text) that APR-P improves performances of both  OOD detection and defense adversarial attacks tasks.
	From Appendix \ref{sec:study}, the adversarial samples are more different from original samples in intermediate and high frequencies, while the OOD samples may share similarities with the original samples in any frequencies.
	The gains of APR-P and APR-S to different frequency domains bring the gains to different tasks.
	
	Furthermore, APR-SP (combining APR-S and APR-P) could substantially improve robustness to most frequency perturbations.
	The weak sensitivity to the intermediate frequencies is reasonable because of the gains for target prediction from intermediate frequencies in Appendix \ref{sec:templates_phase}.

	\subsection{Clean Error} 
	% We include additional results on CIFAR-10. 
	Table \ref{tab:clean} reports clean error \cite{hendrycks2019augmix} of CIFAR-10 by different methods, and the proposed method achieves the best performances on various backbone networks.
	APR-SP not only improves the model adaptability to the common corruptions, surface variations and OOD detection, but also improves the classification accuracy of the clean images.
	
	\begin{table}[!tb]
		\caption{OOD performance of different methods on the larger and more difficult datasets, where ImageNet-1K is the in-distribution dataset and ImageNet-O is the OOD dataset.}
		\centering
		\label{tab:imagenet2}
		\begin{tabular}{cccc}
			% \specialrule{.16em}{0pt} {.65ex}
			\toprule
			Method & AUROC & OSCR \\
			\midrule
			Standard & 40.9 & 36.8 \\
			APR-SP & \textbf{62.3} & \textbf{53.2} \\
			\bottomrule
			% 		\specialrule{.16em}{.4ex}{0pt}
		\end{tabular}
	\end{table}
	
	\subsection{OOD Detection on ImageNet.} 
	We conduct OOD experiments on the larger and more difficult ImageNet-1K dataset \cite{ILSVRC15}. ImageNet-O \cite{hendrycks2019natural} is adopted as the out-of-distribution dataset of ImageNet-1K. ImageNet-O includes 2K examples from ImageNet-22K \cite{ILSVRC15} excluding ImageNet-1K. The ResNet 50 \cite{he2016deep} is trained on ImageNet-1K and tested on both ImageNet-1K and ImageNet-O.
	
	In order to evaluate the accuracy of in-distribution and the ability of OOD detection simultaneously, we introduce \emph{Open Set Classification Rate} (OSCR) \cite{dhamija2018reducing,chen2021adversarial} as an evaluation metric. 
	Let $\delta$ is a score threshold. 
	The \emph{Correct Classification Rate} (CCR) is the fraction of the samples where the correct class $k$ has maximum probability and has a probability greater than $\delta$: 
	\begin{equation}\small
		CCR(\delta)=\frac{|\{x \in \mathcal{D}_\mathcal{I}^k \land arg max_k P(k|x) = \hat{k} \land P(\hat{k}|x) \ge \delta \} }{|\mathcal{D}_\mathcal{I}^k|}.
	\end{equation}
	where $\mathcal{D}_\mathcal{I}^k$ is the interest in-distribution classes that the neural network shall identify. 
	The \emph{False Positive Rate} (FPR) is the fraction of samples from OOD data $\mathcal{D}_\mathcal{O}$ that are classified as \emph{any} in-distribution class $k$ with a probability greater than $\delta$:
	\begin{equation}\small
		FPR(\delta) = \frac{|\{x | x \in \mathcal{D}_\mathcal{O} \land max_k{P(k|x) \ge \delta}\}|}{|\mathcal{D}_\mathcal{O}|}.
	\end{equation}
	A larger value of OSCR indicates a better detection performance. 
	As shown in Table \ref{tab:imagenet2}, APR-SP performs better than the standard augmentations even on the large and difficult dataset. 
	Especially, APR-SP achieves about 22\% improvement on AUROC. 
	From the OSCR, APR-SP improves the performances of the OOD detection while maintaining test accuracy. 
	These results indicate the excellent scalability of APR in larger-scale datasets.

	\subsection{More CAM Visualization Examples}
	
\begin{figure*}[]
		\centering
		\subfigure[Noise: Gaussion]{
			\begin{minipage}[t]{0.48\linewidth}
				\centering
				\includegraphics[width=\linewidth]{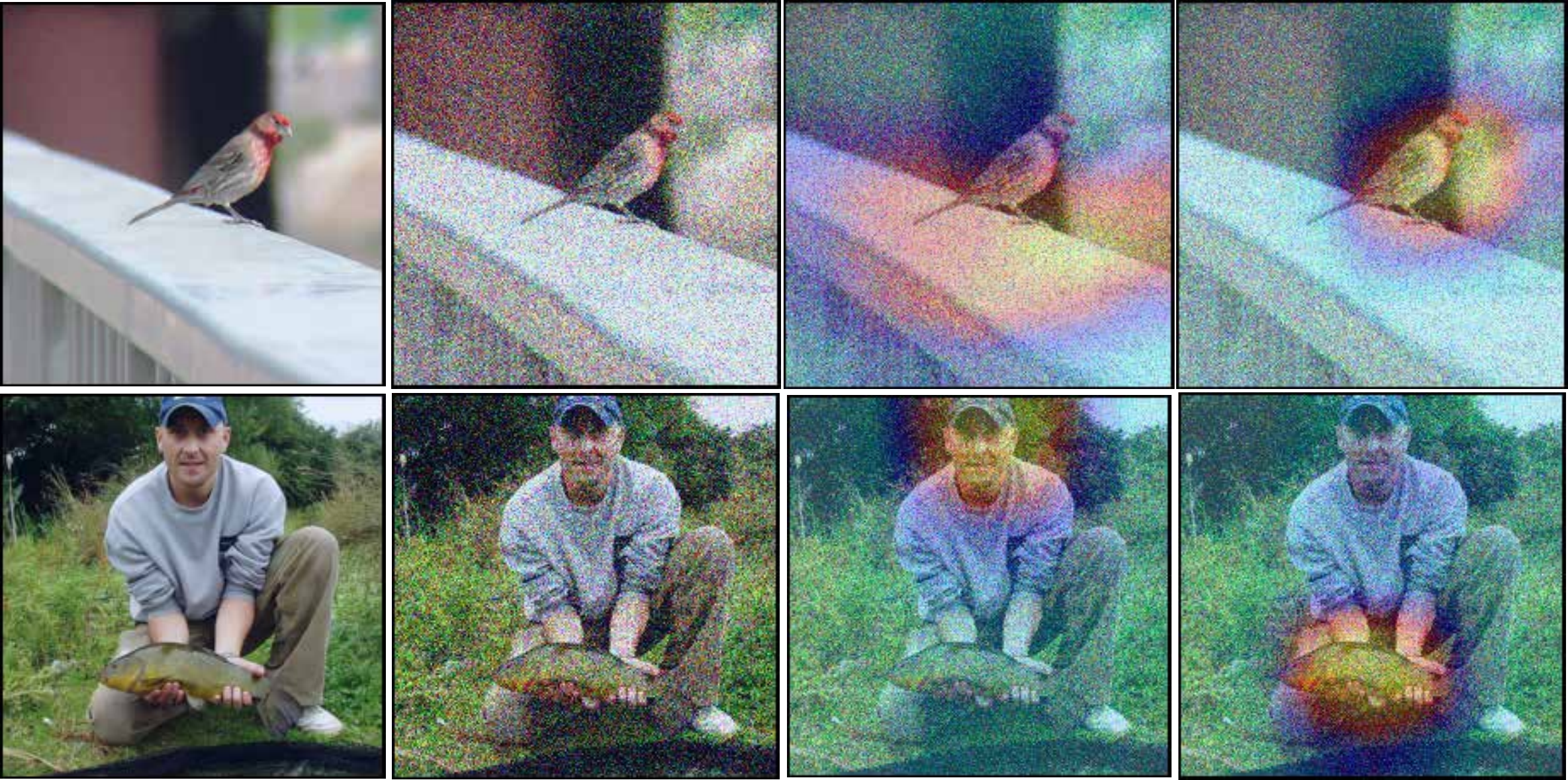}
				\label{fig:gaussion}
			\end{minipage}%
		}
		\subfigure[Noise: Shot]{
			\begin{minipage}[t]{0.48\linewidth}
				\centering
				\includegraphics[width=\linewidth]{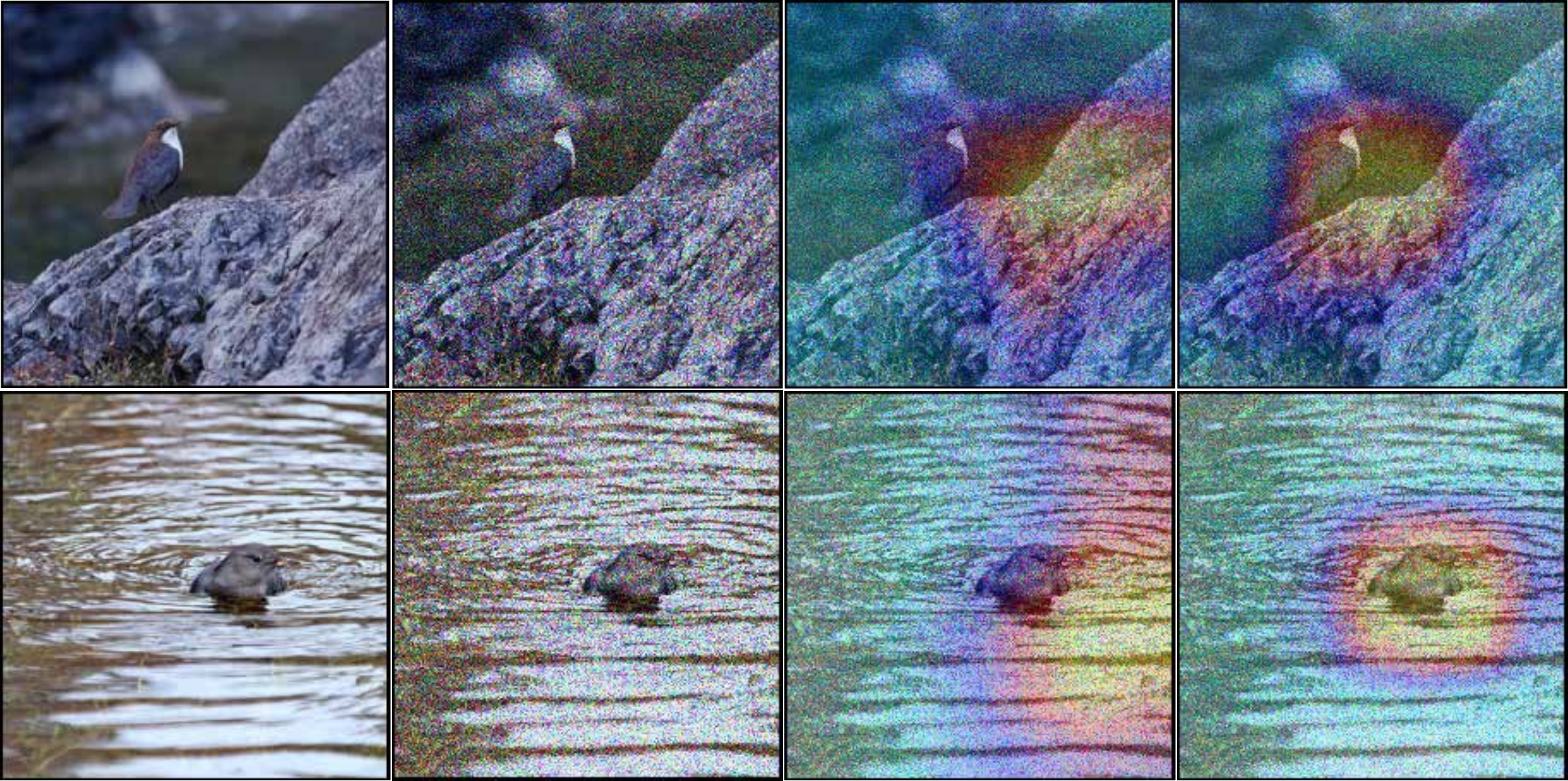}
				\label{fig:shot}
			\end{minipage}%
		}\\
		\subfigure[Noise: Impulse]{
			\begin{minipage}[t]{0.48\linewidth}
				\centering
				\includegraphics[width=\linewidth]{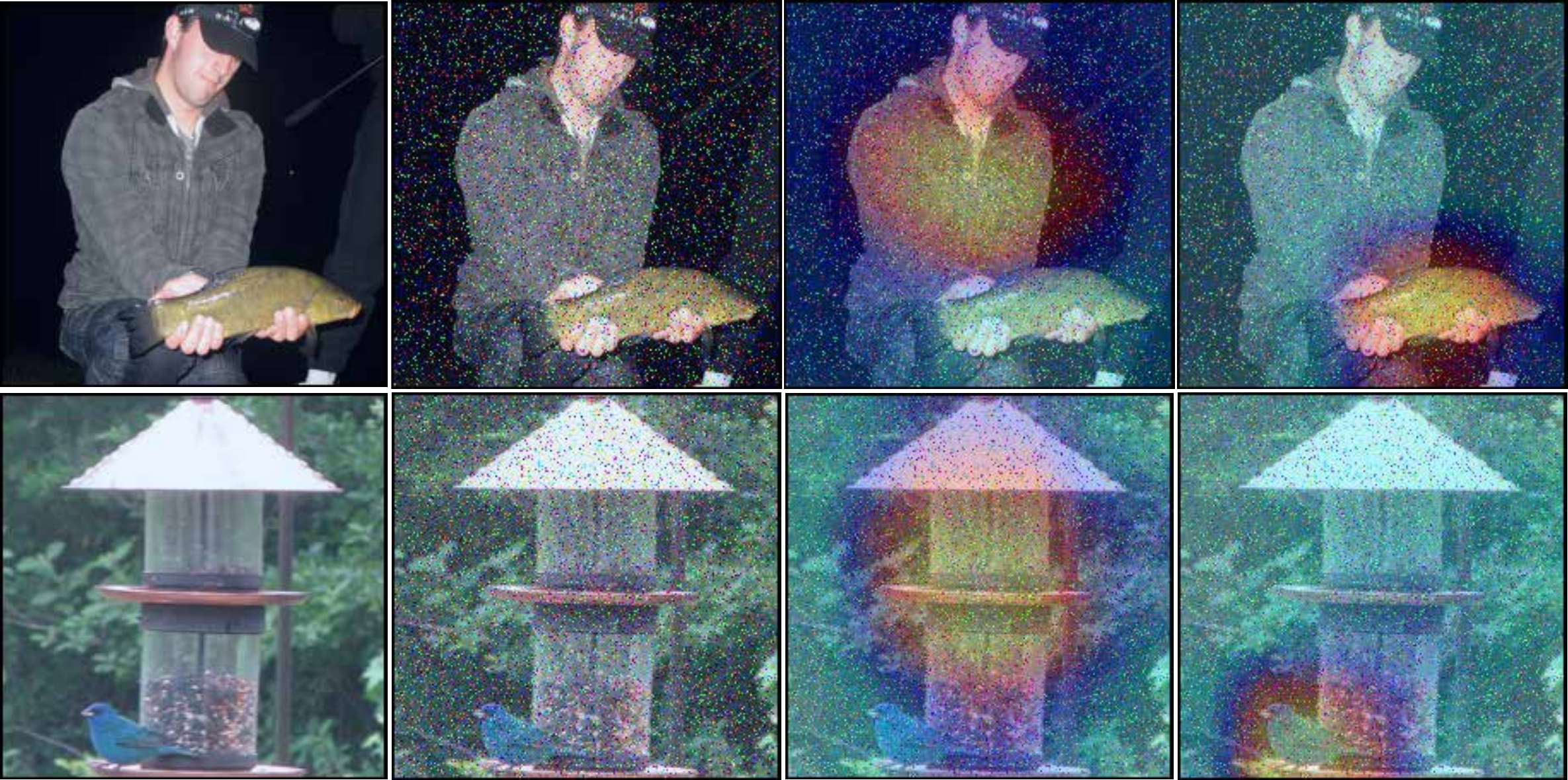}
				\label{fig:impluse}
			\end{minipage}%
		}%
		\subfigure[Blur: Defocus]{
			\begin{minipage}[t]{0.48\linewidth}
				\centering
				\includegraphics[width=\linewidth]{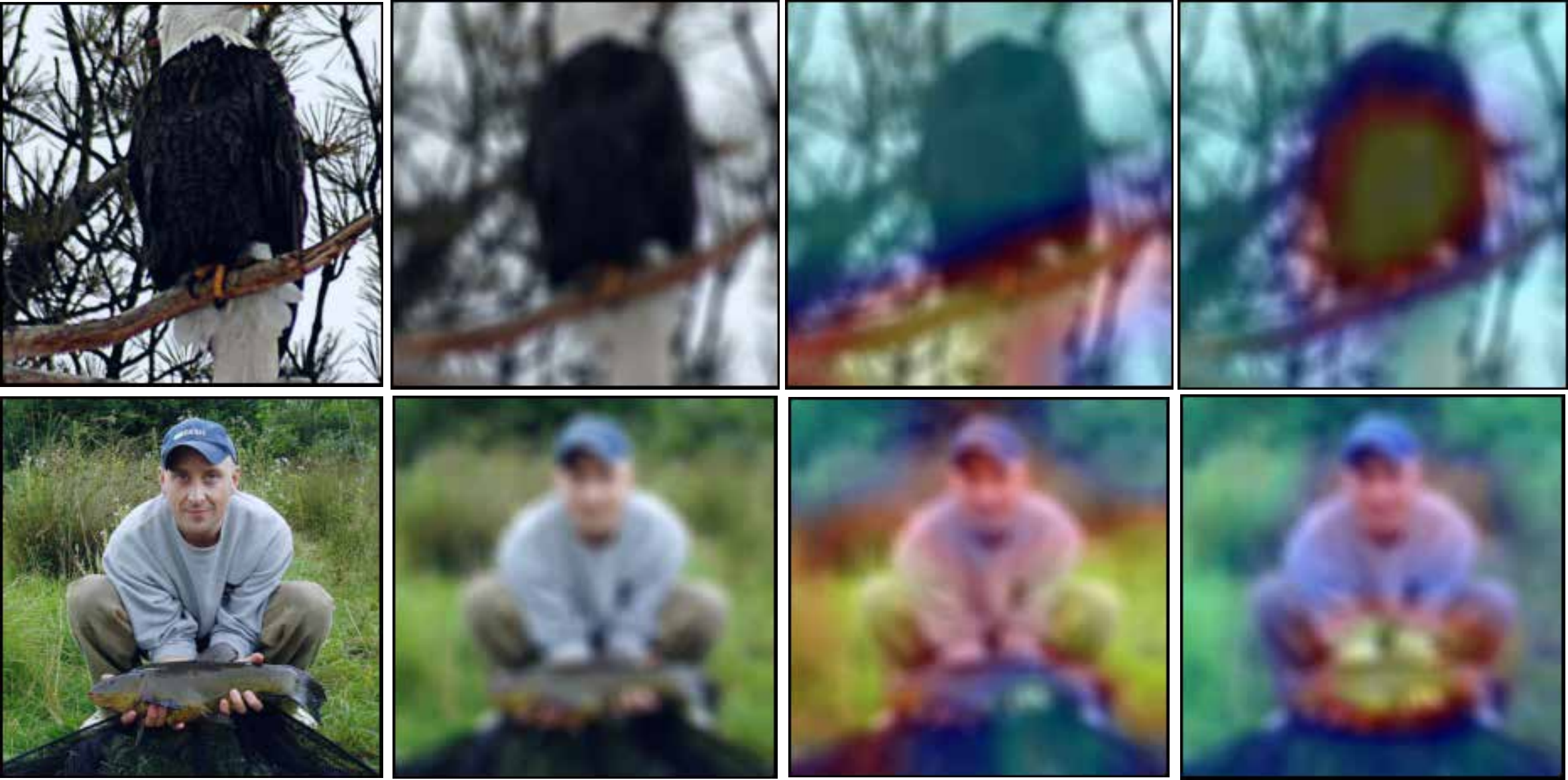}
				\label{fig:defocus}
			\end{minipage}%
		}\\
		\subfigure[Blur: Glass]{
			\begin{minipage}[t]{0.48\linewidth}
				\centering
				\includegraphics[width=\linewidth]{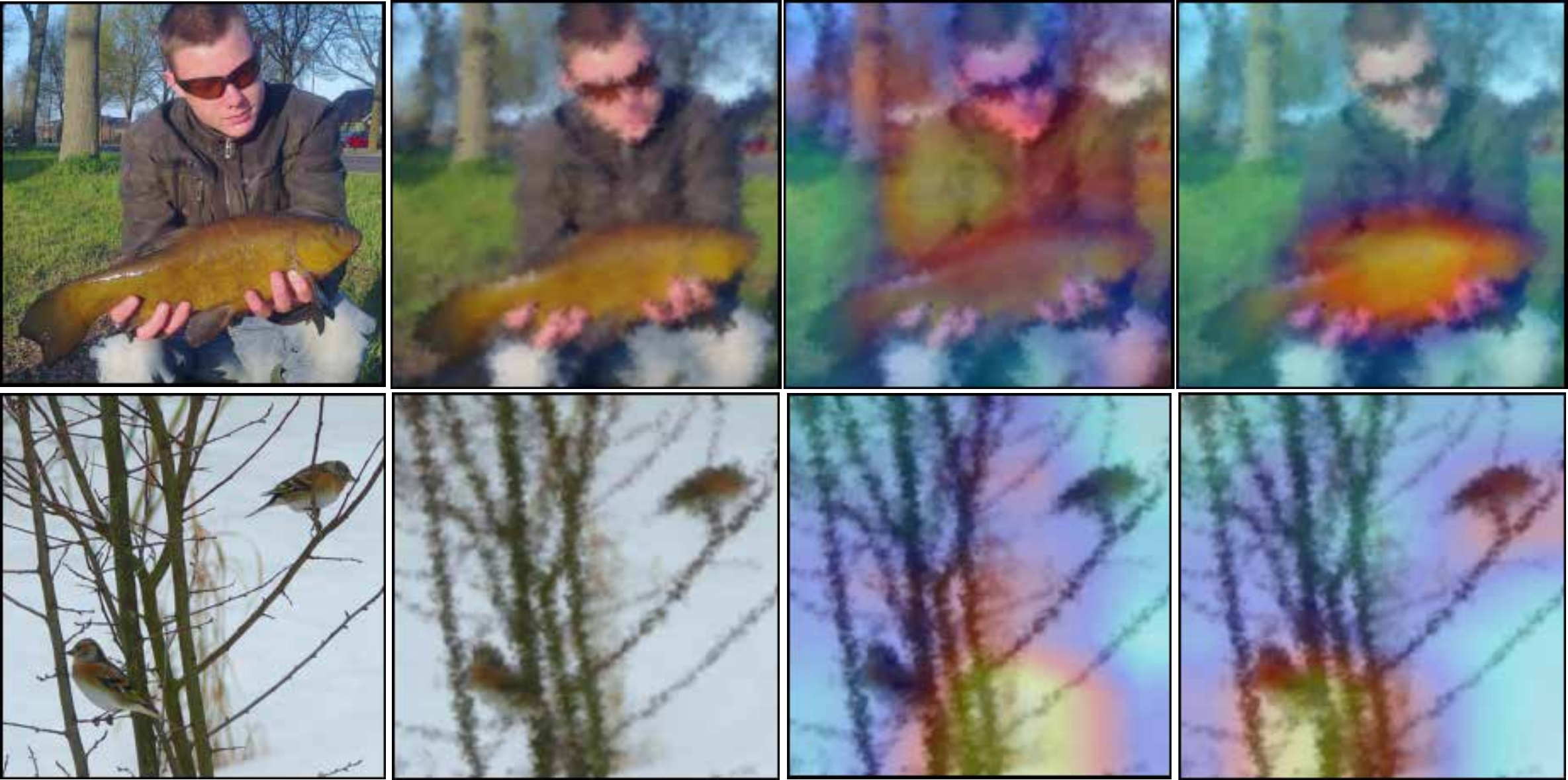}
				\label{fig:glass}
			\end{minipage}%
		}%
		\subfigure[Blur: Motion]{
			\begin{minipage}[t]{0.48\linewidth}
				\centering
				\includegraphics[width=\linewidth]{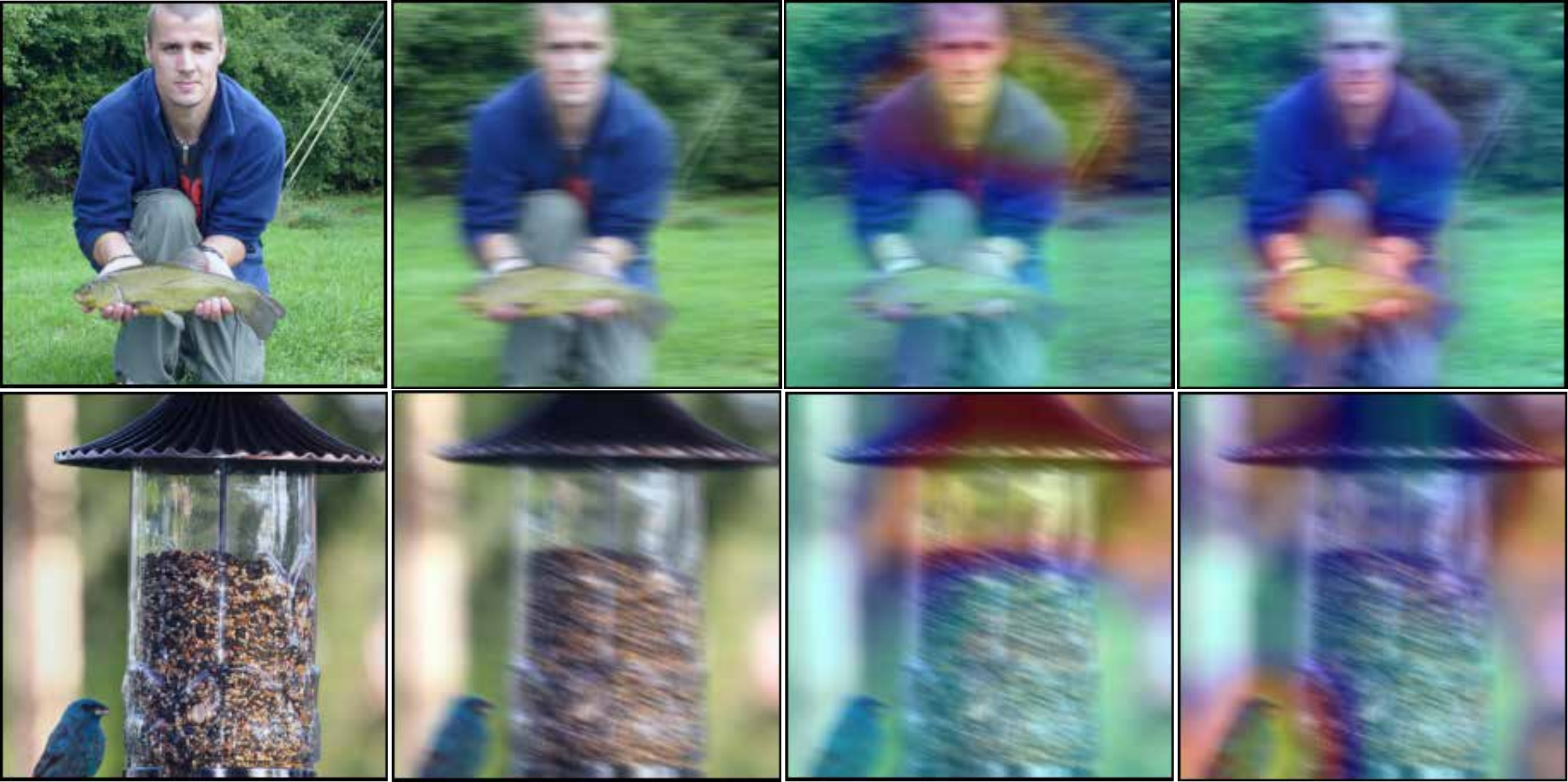}
				\label{fig:motion}
			\end{minipage}%
		}\\
		\subfigure[Blur: Zoom]{
			\begin{minipage}[t]{0.48\linewidth}
				\centering
				\includegraphics[width=\linewidth]{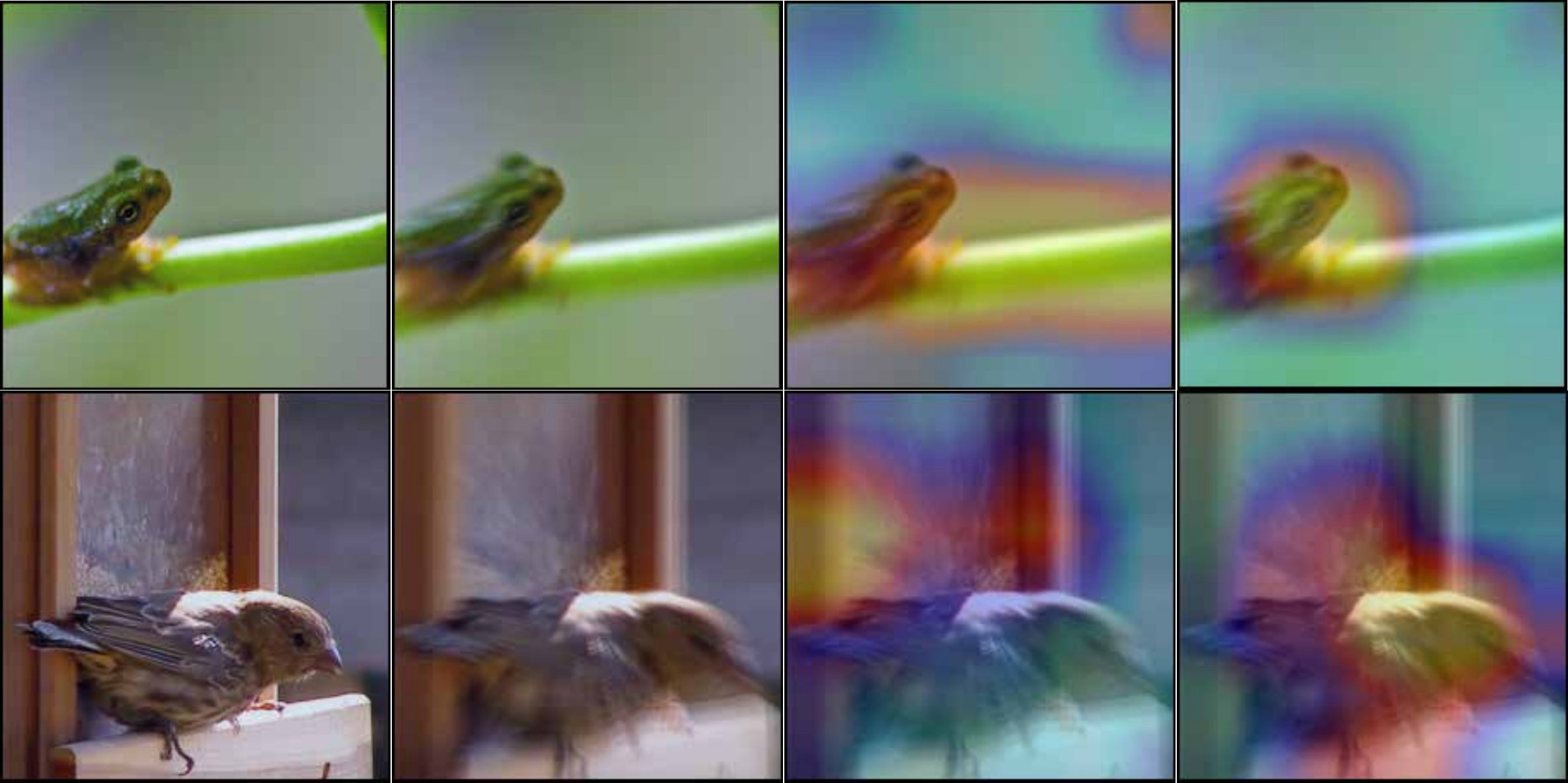}
				\label{fig:zoom}
			\end{minipage}%
		}%
		\subfigure[Weather: Snow]{
			\begin{minipage}[t]{0.48\linewidth}
				\centering
				\includegraphics[width=\linewidth]{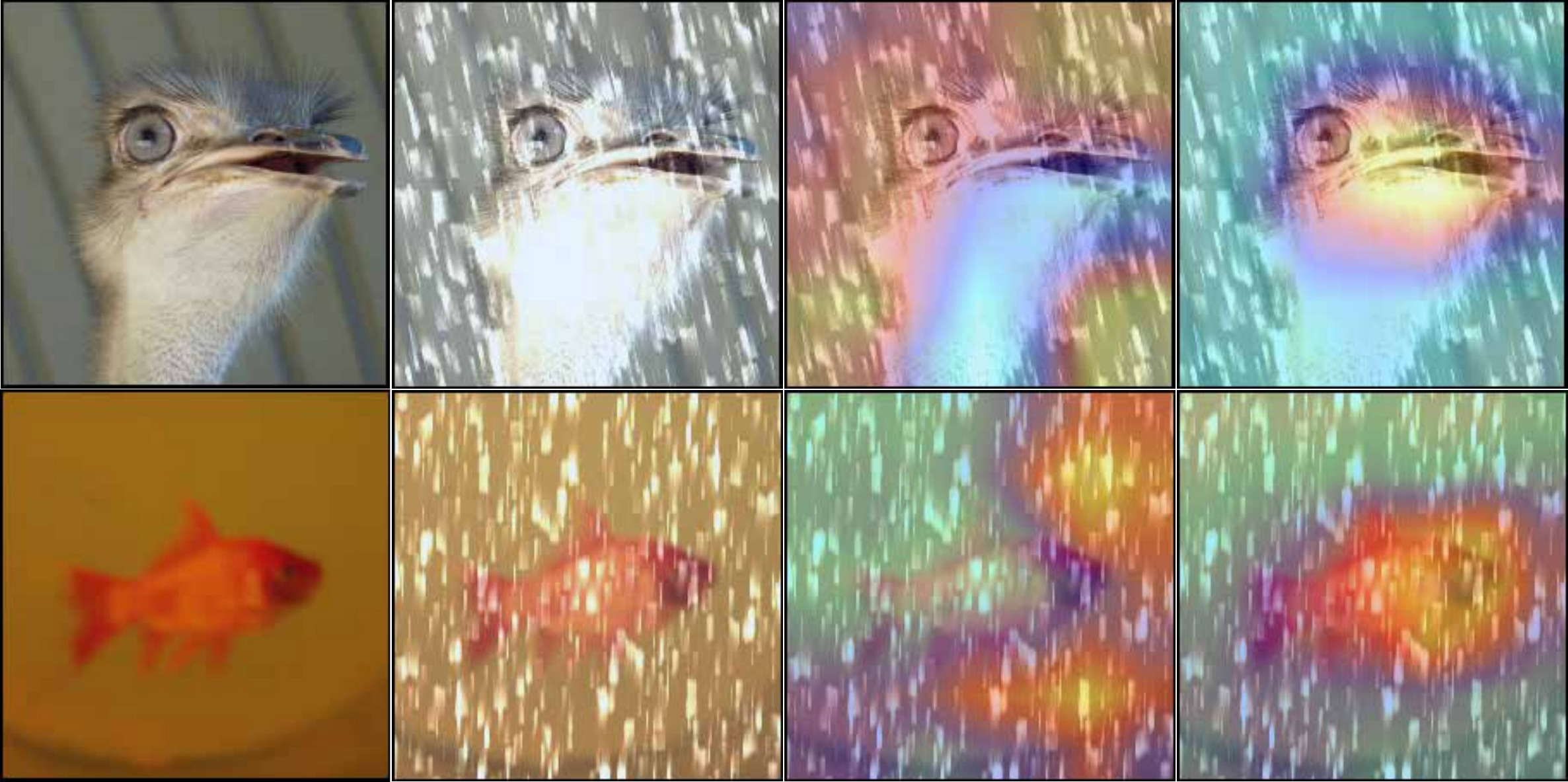}
				\label{fig:snow}
			\end{minipage}%
		}
		\caption{The Gradient-weighted Class Activation Mapping \cite{selvaraju2017grad} of the baseline (the third column in each panel) and the proposed APR-SP (the fourth column in each panel) for images with different common corruptions and surface variations (the second column in each panel). The original images are in the first column in each panel. Best viewed in color. APR-SP still is robust even in various corruptions.}
		\label{fig:corr1}
	\end{figure*}

\begin{figure*}[]
		\centering
		\subfigure[Weather: Fog]{
			\begin{minipage}[t]{0.48\linewidth}
				\centering
				\includegraphics[width=\linewidth]{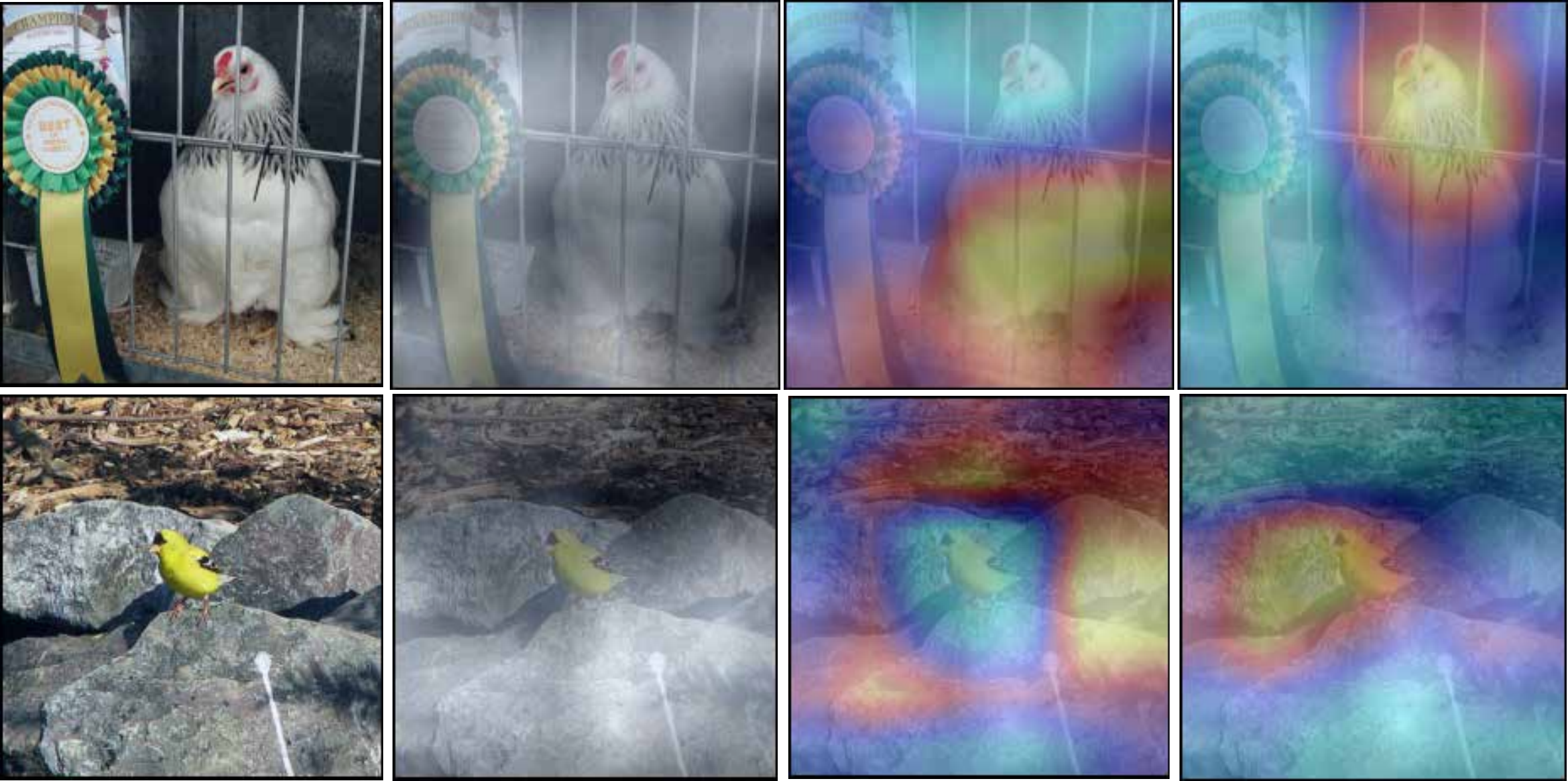}
				\label{fig:fog}
			\end{minipage}%
		}%
		\subfigure[Digital: JPEG Compression]{
			\begin{minipage}[t]{0.48\linewidth}
				\centering
				\includegraphics[width=\linewidth]{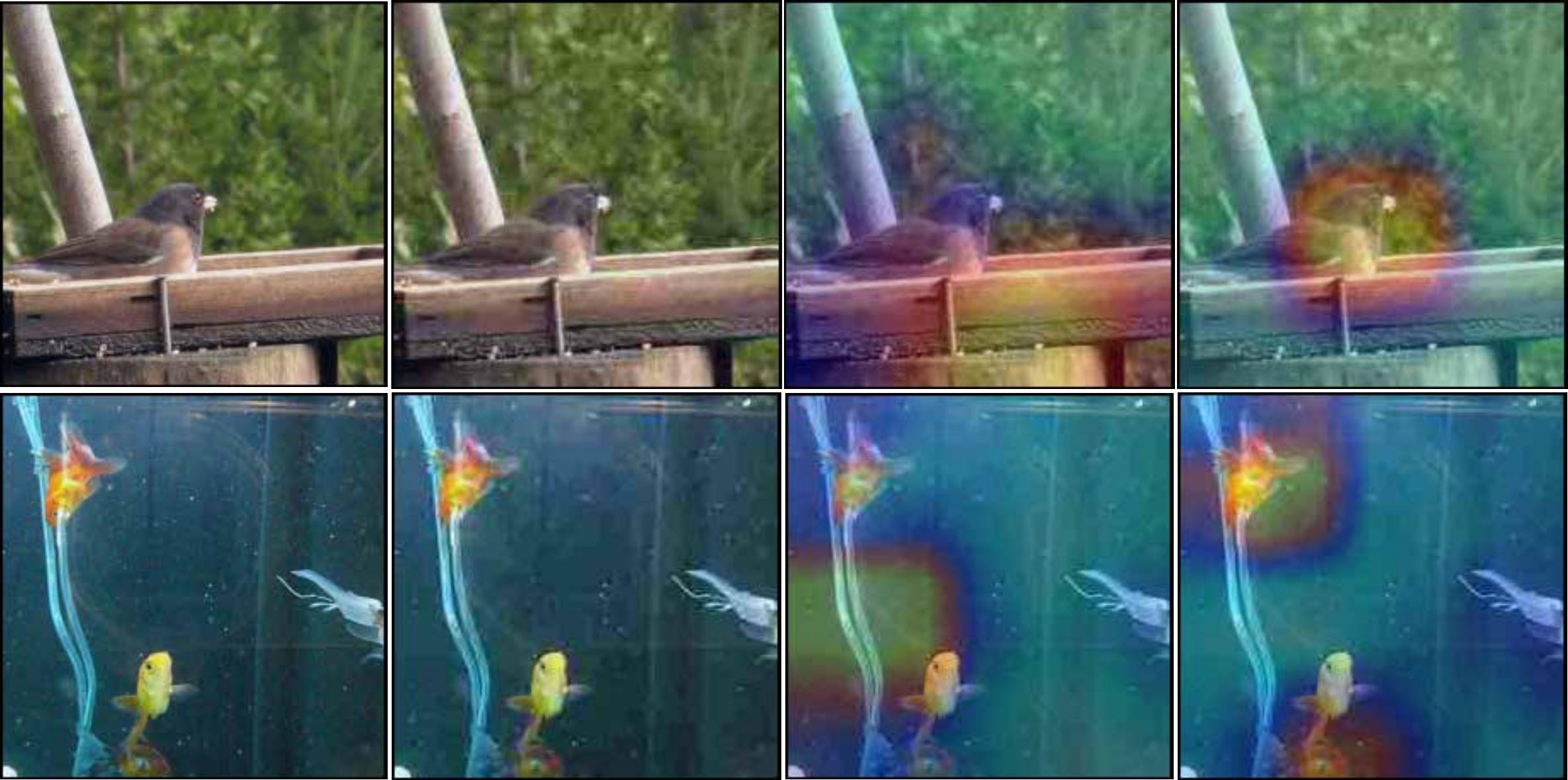}
				\label{fig:jpeg}
			\end{minipage}%
		}\\
		\subfigure[Digital: Elastic Transform]{
			\begin{minipage}[t]{0.48\linewidth}
				\centering
				\includegraphics[width=\linewidth]{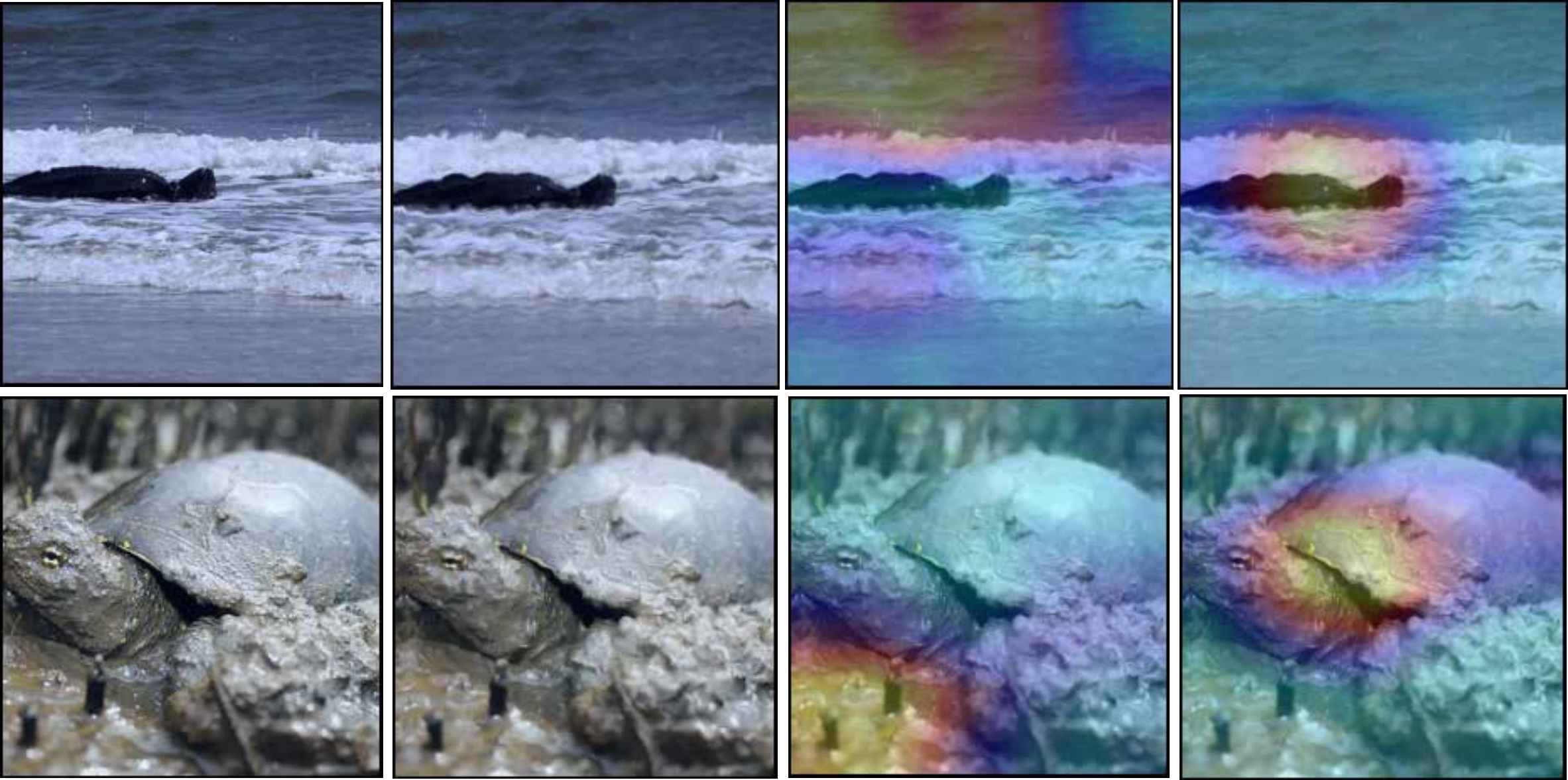}
				\label{fig:elastic}
			\end{minipage}%
		}%
		\subfigure[Digital: Pixelate]{
			\begin{minipage}[t]{0.48\linewidth}
				\centering
				\includegraphics[width=\linewidth]{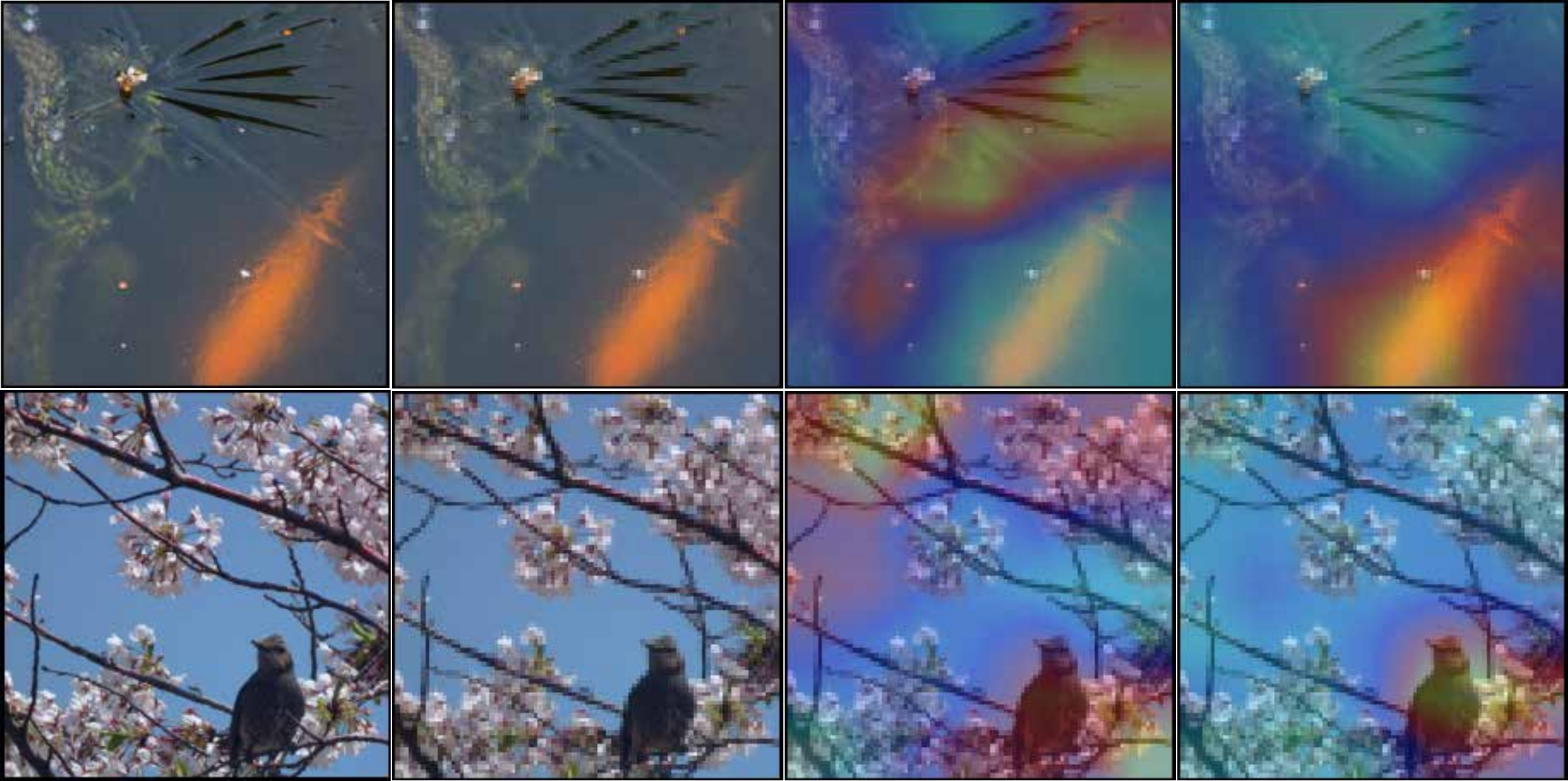}
				\label{fig:pixelate}
			\end{minipage}%
		}%
		\caption{The Gradient-weighted Class Activation Mapping \cite{selvaraju2017grad} of the baseline (the third column in each panel) and the proposed APR-SP (the fourth column in each panel) for images with different common corruptions and surface variations (the second column in each panel). The original images are in the first column in each panel. Best viewed in color. APR-SP still is robust even in various corruptions.}
		\label{fig:corr2}
	\end{figure*}

	We also list more visualization examples with various corruptions in Figure \ref{fig:corr1} and \ref{fig:corr2}, the CNN trained by APR-SP is able to focus on the target objects for classification even with different common corruptions and surface variations.
	
\end{document}